\documentclass[final]{cvpr}

\usepackage{times}
\usepackage{epsfig}
\usepackage{graphicx}
\usepackage{amsmath}
\usepackage{amssymb}

\usepackage[pagebackref=true,breaklinks=true,colorlinks,bookmarks=false]{hyperref}

\usepackage{color}
\usepackage{graphicx}
\usepackage{amsmath}
\usepackage{caption}
\usepackage{subcaption}

\title{PCLs: Geometry-aware Neural Reconstruction of 3D Pose\\with Perspective Crop Layers}

\author{
	Frank Yu$^1$
	\and
	Mathieu Salzmann$^2$
	\and
	Pascal Fua$^2$
\and
	Helge Rhodin$^{1}$
	\and
	\\
	$^1$UBC, Vancouver, Canada\\
	$^2$EPFL, Lausanne, Switzerland\\
	
	{\tt\small \{frankyu, rhodin\}@cs.ubc.ca}
}

\begin{document}

\maketitle

\definecolor{olive}{RGB}{50,150,50}
\definecolor{frank}{RGB}{198, 3, 252}

\newif\ifdraft
\drafttrue

\ifdraft
 \newcommand{\PF}[1]{{\color{red}{\bf pf: #1}}}
 \newcommand{\pf}[1]{{\color{red} #1}}
 \newcommand{\HR}[1]{{\color{blue}{\bf hr: #1}}}
 \newcommand{\hr}[1]{{\color{blue} #1}}
 \newcommand{\VC}[1]{{\color{green}{\bf vc: #1}}}
  \newcommand{\vc}[1]{{\color{green} #1}}
 \newcommand{\ms}[1]{{\color{olive}{#1}}}
 \newcommand{\MS}[1]{{\color{olive}{\bf ms: #1}}}
 \newcommand{\JS}[1]{{\color{cyan}{\bf js: #1}}}
 \newcommand{\fy}[1]{{\color{frank}{#1}}}
 \newcommand{\FY}[1]{{\color{frank}{\bf fy: #1}}}
  \newcommand{\jl}[1]{{\color{cyan}{#1}}}
 \newcommand{\JL}[1]{{\color{cyan}{\bf jl: #1}}}

\else
 \newcommand{\PF}[1]{{\color{red}{}	
 \newcommand{\pf}[1]{ #1 }
 \newcommand{\HR}[1]{{\color{blue}{}
 \newcommand{\hr}[1]{ #1 }
 \newcommand{\VC}[1]{{\color{green}{}
 \newcommand{\ms}[1]{ #1 }
 \newcommand{\MS}[1]{{\color{olive}{}
\fi

\newcommand{\TODO}[1]{\textcolor{red}{TODO: #1}}
\newcommand{\todo}[1]{\textcolor{red}{#1}}
\newcommand{\R}{\mathbb{R}}
\newcommand{\fr}{t}
\newcommand{\pcentroid}{\hat{p}}

\newcommand{\supp}{appendix}

\newcommand{\mA}{\mathbf{A}}
\newcommand{\mB}{\mathbf{B}}
\newcommand{\mC}{\mathbf{C}}
\newcommand{\mD}{\mathbf{D}}
\newcommand{\mE}{\mathbf{E}}
\newcommand{\mF}{\mathbf{F}}
\newcommand{\mG}{\mathbf{G}}
\newcommand{\mGamma}{\mathbf{\Gamma}}
\newcommand{\mH}{\mathbf{H}}
\newcommand{\mI}{\mathbf{I}}
\newcommand{\mJ}{\mathbf{J}}
\newcommand{\mK}{\mathbf{K}}
\newcommand{\mL}{\mathbf{L}}
\newcommand{\mM}{\mathbf{M}}
\newcommand{\mN}{\mathbf{N}}
\newcommand{\mO}{\mathbf{O}}
\newcommand{\mP}{\mathbf{P}}
\newcommand{\mQ}{\mathbf{Q}}
\newcommand{\mR}{\mathbf{R}}
\newcommand{\mRvr}{\mathbf{R}_{\text{virt}\rightarrow\text{real}}}
\newcommand{\mRrv}{\mRvr^{-1}}%
\newcommand{\mS}{\mathbf{S}}
\newcommand{\mT}{\mathbf{T}}
\newcommand{\mU}{\mathbf{U}}
\newcommand{\mV}{\mathbf{V}}
\newcommand{\mW}{\mathbf{W}}
\newcommand{\mX}{\mathbf{X}}
\newcommand{\mY}{\mathbf{Y}}
\newcommand{\mZ}{\mathbf{Z}}

\newcommand{\cA}{\mathcal A}
\newcommand{\cB}{\mathcal B}
\newcommand{\cC}{\mathcal C}
\newcommand{\cD}{\mathcal D}
\newcommand{\cE}{\mathcal E}
\newcommand{\cF}{\mathcal F}
\newcommand{\cG}{\mathcal G}
\newcommand{\cH}{\mathcal H}
\newcommand{\cI}{\mathcal I}
\newcommand{\cJ}{\mathcal J}
\newcommand{\cK}{\mathcal K}
\newcommand{\cL}{\mathcal L}
\newcommand{\cM}{\mathcal M}
\newcommand{\cN}{\mathcal N}
\newcommand{\cO}{\mathcal O}
\newcommand{\cP}{\mathcal P}
\newcommand{\cQ}{\mathcal Q}
\newcommand{\cR}{\mathcal R}
\newcommand{\cS}{\mathcal S}
\newcommand{\cT}{\mathcal T}
\newcommand{\cU}{\mathcal U}
\newcommand{\cV}{\mathcal V}
\newcommand{\cW}{\mathcal W}
\newcommand{\cX}{\mathcal X}
\newcommand{\cY}{\mathcal Y}
\newcommand{\cZ}{\mathcal Z}

\newcommand{\va}{\mathbf{a}}
\newcommand{\vb}{\mathbf{b}}
\renewcommand{\vc}{\mathbf{c}}
\newcommand{\vd}{\mathbf{d}}
\newcommand{\ve}{\mathbf{e}}
\newcommand{\vf}{\mathbf{f}}
\newcommand{\vg}{\mathbf{g}}
\newcommand{\vh}{\mathbf{h}}
\newcommand{\vi}{\mathbf{i}}
\newcommand{\vj}{\mathbf{j}}
\newcommand{\vk}{\mathbf{k}}
\newcommand{\vl}{\mathbf{l}}
\newcommand{\vm}{\mathbf{m}}
\newcommand{\vn}{\mathbf{n}}
\newcommand{\vo}{\mathbf{o}}
\newcommand{\vp}{\mathbf{p}}
\newcommand{\vq}{\mathbf{q}}
\newcommand{\vr}{\mathbf{r}}
\renewcommand{\vs}{\mathbf{s}}
\newcommand{\vt}{\mathbf{t}}
\newcommand{\vu}{\mathbf{u}}
\newcommand{\vv}{\mathbf{v}}
\newcommand{\vw}{\mathbf{w}}
\newcommand{\vx}{\mathbf{x}}
\newcommand{\vy}{\mathbf{y}}
\newcommand{\vz}{\mathbf{z}}

\newcommand{\comment}[1]{}

\newcommand{\preal}{\mathbf{p}}
\newcommand{\pvirt}{\mathbf{p}_\text{virt}}
\newcommand{\vreal}{\mathbf{v}}
\newcommand{\vvirt}{\mathbf{v}_\text{virt}}
\newcommand{\Kreal}{\mathbf{K}}
\newcommand{\Kvirt}{\mathbf{K}^\text{virt}}
\newcommand{\Kcan}{\mathbf{K}_\text{canonical}}

\newcommand{\argmin}{\operatornamewithlimits{argmin}}

\begin{abstract}

Local processing is an essential feature of CNNs and other neural network architectures---it is one of the reasons why they work so well on images where relevant information is, to a large extent, local.
However, perspective effects stemming from the projection in a conventional camera vary for different global positions in the image. %
We introduce Perspective Crop Layers (PCLs)---a form of perspective crop of the region of interest based on the camera geometry--- and show that accounting for the perspective consistently improves the accuracy of state-of-the-art 3D pose reconstruction methods.
PCLs are modular neural network layers, which, when inserted into existing CNN and MLP architectures, deterministically remove the location-dependent perspective effects while leaving end-to-end training and the number of parameters of the underlying neural network unchanged.
We demonstrate that PCL leads to improved 3D human pose reconstruction accuracy for CNN architectures that use cropping operations, such as spatial transformer networks (STN), and, somewhat surprisingly, MLPs used for 2D-to-3D keypoint lifting. 
Our conclusion is that it is important to utilize camera calibration information when available, for classical and deep-learning-based computer vision alike. 
PCL offers an easy way to improve the accuracy of existing 3D reconstruction networks by making them geometry-aware.
Our code is publicly available at \href{https://github.com/yu-frank/PerspectiveCropLayers}{github.com/yu-frank/PerspectiveCropLayers}.

\end{abstract}

\section{Introduction} 
Convolutional neural networks (CNNs) have proven highly effective for image-based prediction tasks because of their translation invariance and the locality of the computation they perform. For 3D pose estimation, this allows them to focus on image locations that carry information about the pose while discarding other ones \cite{Tekin2016BMVC,Park2016,Du2016,VNect_SIGGRAPH2017,Pavlakos2017,lcrnet2017,Tome_2017_CVPR,sun2017compositional,OriNet2018,zhou2019hemlets,Li_2020_CVPR,Xu_2020_CVPR}.%

\comment{
\fy{The field of 3D human pose estimation has seen tremendous growth and alongside the developments of deeper and more expressive neural networks (NN). Recent approaches tackle this problem utilize convolutional neural networks (CNNs) which have been proven highly effective for image-based prediction tasks because of their translation invariance. \TODO{cite~\cite{}} Others, offer a different path my using multilayer perceptrons (MLPs) to lift 2D keypoints into 3D keypoints \TODO{cite~\cite{}}. However, while these approaches have achieved great results on numerous benchmarks, they explicitly ignore the perspective distortions that come with projecting a three dimensional object (ie. people) to a flat two dimensional image plane.}
}

 \begin{figure}[t!]
\begin{center}
	\includegraphics[width=1\linewidth,trim={8.5cm 2.1cm 8.5cm 2.1cm},clip]{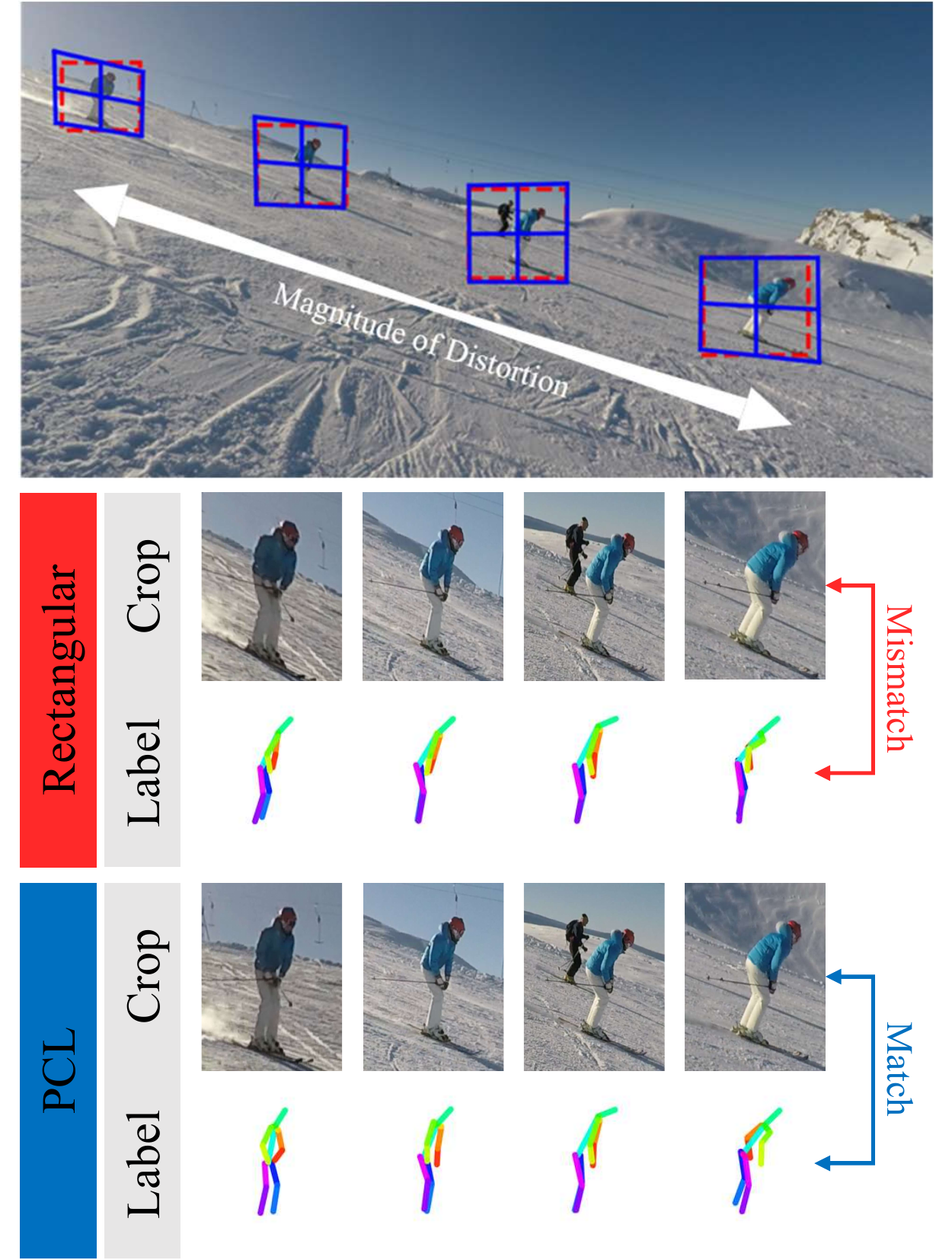}
\end{center}
\caption{\label{fig:teaser}
		{\bf Perspective effects and correction with PCL crops.} 
		The skier looks as if she turns, from front to backwards facing, although recorded with a static camera and going straight. PCLs correct the stretching originating from the projection onto the image plane and matches the 3D pose label to the local view direction of the crop. 
	}
\end{figure}
Convolutions in the image plane, however, ignore the perspective effects caused by projecting a 3D scene in 2D. For example, as shown in Figure~\ref{fig:teaser}, a person captured by static camera in a fixed pose and moving in a constant direction is seen from different angles as their image location changes. Applying the same convolutional filter at the top-left image corner and at the bottom-right one will therefore yield different features, even though the pose is the same.  In practice, this is typically tackled by increasing the width and depth of the network, so that different filters and layers can model the same 3D pose perspective-distorted in different ways. The effectiveness of this procedure, however, strongly depends on the availability of large amounts of training data, which is far from being a given for pose estimation in the wild. Notably, two-stage approaches that lift 3D pose from 2D pose estimates using multilayer perceptrons (MLPs)~\cite{chen20173d,Moreno_cvpr2017,fang2017learning,mpii3Dhp2017,martinez_2017_3Dbaseline,Hossain2018,pavlakos2018ordinal,XNect_SIGGRAPH2020} rely also on translational invariance by centering the 2D pose on a root joint, thereby losing important cues on perspective distortion too.

In this paper, we therefore introduce Perspective Crop Layers (PCLs) to explicitly account for perspective distortion within CNNs and other neural networks. Specifically, we use a homography to map the input image to a virtual camera with pre-defined intrinsic parameters that point to the region of interest (RoI). The homography parameters are functions of the RoI's location and scale. Hence, this yields a synthetic view in which the location-dependent perspective deformations are undone. The 3D pose inferred from this synthetic view  can then be projected into the original image.  This requires {\it a priori} knowledge about the intrinsic camera parameters, which is rarely a problem in real-world situations because they either are readily available from the camera specifications or can be inferred from the input images alone~\cite{Workman15,Hold18}. We will further show that our PCLs are robust to calibration inaccuracies. Ultimately, all the operations performed by our PCLs are differentiable, and thus amenable to end-to-end learning, while removing the need for the CNN to learn the already known perspective geometry.
Our contributions can be summarized as follows:
\begin{itemize}
\item We showcase the influence of perspective effects on 3D pose estimates that increases for poses away from the image center, which is disregarded by virtually all state-of-the-art algorithms;
\item We derive the equations to compensate for these effects across the image in a location-dependent manner;
\item We encapsulate our formalism into generic NN layers, dubbed PCLs, that naturally integrate into existing deep learning frameworks.
\end{itemize}
We demonstrate the benefits of our PCLs for 3D human pose estimation of both rigid objects and articulated people. PCLs yield a consistent boost in performance, of $2-10\%$ on average and up to $25\%$ at the image boundary where perspective effects are strongest.
Notably, the improvements attributable to our PCLs are consistent across the baseline we seek to improve, which validates our claim that even the most-advance deep networks do not learn these perspective effects on the existing datasets. This includes a PCL variant that undoes the perspective effect on 2D keypoints, thus allowing us to showcase the benefits of our approach on state-of-the-art 3D pose estimation methods that lift 2D keypoint detections to 3D poses \cite{martinez_2017_3Dbaseline,pavllo20193d}. %
Our code is publicly available at \href{https://github.com/yu-frank/PerspectiveCropLayers}{github.com/yu-frank/PerspectiveCropLayers}.

\comment{
and the geometric modeling of the virtual camera enables us to transform locally-inferred 3D pose back to the original, global camera frame. This explicit modeling frees the neural network from the need to learn to account for perspective effects. Fig.~\ref{fig:error map} shows how this improves reconstruction errors on the popular Human 3.6 million benchmark, as a function of the person position in the image. \JL{h36m needs citation}

Notably, the improvement we observe due to PCL increases with the reconstruction accuracy of the baseline since other error source dominate for less accurate networks \TODO{double check if we can show this}, demonstrating that even deep networks are not good at learning perspective effects and making PCL an even more important ingredient for the state-of-the-art algorithms. Moreover, we demonstrate that the defacto standard strategy for state-of-the-art 3D pose estimation methods of lifting 2D keypoint detections with a multilayer perceptron (MLP) to 3D pose suffers from the same locality issues as CNN-based architectures since the 2D pose is usually centered at the pelvis joint to avoid overfitting to global position. We therefore propose a PCL variant that removes the perspective effect on keypoints.

In practice, the synthetic crops of the RoI that PCLs produce are related to the original image by a homography whose parameters are a function of its location and scale. It is differentiable and does not require additional neural network parameters to be learned. This preserves the translation, scale, and rotation 
equivariance of convolutions and affine spatial transformers while undistorting the RoI so that the result becomes location-invariant. Moreover, the explicit modeling of a virtual camera lets us map locally inferred reconstructions, such as articulated 3D human pose, back to the global coordinates of the original camera, as shown in Fig.~\ref{fig:overview}. This requires {\it a priori} knowledge of the camera parameters, which is rarely a problem in real-world situations because they either are readily available from the camera specifications, can be inferred from the input images alone~\cite{Workman15,Hold18}, and, as we show, can be approximate since PCL is robust against calibration inaccuracies.

The geometric relations for inferring the perspective compensation in PCL are not new. Our contributions are
\begin{itemize}
\item deriving the equations to compensate the rotation, foreshortening, and scale of objects in relation to their position on the image plane,  
\item showcasing that the perspective effects that are disregarded by virtually all state-of-the-art 3D pose estimation algorithms do influence 3D pose estimation significantly,
\item demonstrating that 2D to 3D lifting approaches suffer from the same limitations as CNN-based methods and
\item carefully encapsulating PCLs as generic NN layers that naturally integrate into existing deep learning frameworks for which we will release the source code. %
\end{itemize}

We evaluate our PCLs for the tasks of 3D pose estimation of rigid objects and articulated human pose. This is demonstrated by a  boost in performance for 3D pose estimation, both for CNN architectures as well as MLPs lifting 2D pose to 3D pose.

}

\comment{

While shallow, fully-connected networks are universal approximators, specialized deep networks are most successful in practice. This is in part because hand-crafted layers can exploit inherent domain-specific structure. Concerning neural image processing, CNNs are indispensable for exploiting the regular pixel grid and feature locality in natural images. Moreover, recent architectures utilize region of interest pooling and cropping with spatial transformer layers \cite{Girshick15,He17,Dai16,Jaderberg15}, to succeed even if the object of interest covers only a small part of the image. 

Although these approaches exploit the 2D structure of natural images well, they neglect the image formation from a 3D scene to the flat 2D image.
In particular, the effect of image location-dependent perspective projection is widely overlooked. The significant of this effect is visualized in Figure~\ref{fig:teaser}. Albeit not making turns, the pictured skier stretches and changes view angle in relation to the projection direction, which varies across the image location. 

In principle, a neural network could learn such location dependency from examples, however, local processing with attention windows and translation-invariant convolutions hides the location information needed to infer the spatial projection direction; the cause of the perspective effect.

In this paper, we advocate a perspective crop layer (PCL) that treat cropping and local processing as a change of camera perspective. We deterministically map the input image to a virtual camera that points to the region of interest and has pre-defined intrinsic parameters. In this view, location dependent perspective deformations are undone, and the geometric modeling of the virtual camera allows us to transform locally inferred properties back to the original, global camera frame. This deterministic, explicit modeling frees the neural network from the difficulty of learning the perspective effect, leading to improved reconstructions.

}
\comment{
	dominate 3D computer vision methods, because of their 
	learn the mapping from image to 3D representation from examples.
	For instance, for motion capture the location of the human body joints are predicted from an input video.
	To improve training with an attention mechanism, it is common to normalize observations by cropping and re-scaling of the object of interest in the input image before starting the actual 3D reconstruction. 
	The effect is two-fold. 
	First, the reduced image size for this attention window improves computational efficiency. Second, the rescaling mitigates scale variations that have to be learned for 3D prediction.
	
	Naive cropping, however, leads to artifacts. We propose two cropping approaches that respect 3D geometry and projective effects.
	They can be used as drop-in replacement with any existing network architecture. Our evaluations reveal significant and consistent improvement across a multitude of established benchmarks.
	
	The proposed perspective cropping requires knowledge of the camera intrinsic parameters. 
	We utilize the relation between crop location, 3D pose and intrinsic parameters to calibrate unknown cameras intrinsically and extrinsically in a fully automatic way. We demonstrate that humans, although dynamic in motion, are sufficient to accurately infer the complete set of camera parameters of a pinhole projection model. 
	Any human subject is turned into an accurate calibration object by formulating and exploiting the dependency of 3D pose estimates on the intrinsic camera parameters. No calibration pattern or known object is required to be present. 
	
	This calibration approach is applicable to any scenario where intrinsic and extrinsic parameters are needed and human subjects are present, hence, can be used as a calibration tool beyond human motion capture.
}

\section{Related Work}

In this section, we discuss existing ways of handling image distortions and review the existing attention window mechanisms upon which PCLs are built.

\paragraph{Handling perspective effects. }
Many works sidestep perspective effects by training and testing on synthetic renderings \cite{Chang15,Fan17a,Yang15weakly,Yan16} or real images \cite{Hinton11,Yang15weakly} where the object of interest is centered manually. However, these methods are not applicable to natural images where the object can be at an arbitrary location.
If the object location is known in advance, perspective distortion
can be undone in a preprocessing stage. For instance,~\cite{Mehta17a} propose to rotate locally inferred 3D poses back to the camera frame. This strategy has later been adopted by~\cite{Kanazawa18}, 
but neither of these works undistorts the input images or input 2D pose. \cite{Rhodin18a,Rhodin18b} apply an image correction, however, only approximating the homography with an affine transformation.
In other words, the above-mentioned approaches neither model the perspective correction geometrically accurately nor formulate it as a differentiable layer. However, differentiability is an important prerequisite for end-to-end training on natural images, particularly for unsupervised approaches, that deal with unknown object locations.

Related are classical methods that undo the Keystone effect, wherein distortion is caused when an image is projected to a non-perpendicular wall \cite{keystone2000}. The difficulty lies in deriving the deformation parameters, which we derive analytically for our setting while keystoning requires fitting.

\paragraph{Radial undistortion.}
Fisheye cameras and others with a large field of view yield large deformations when mapped to a rectangular pixel grid. 
They are better represented with spherical images, thereby avoiding location-dependent deformation entirely.
This, however, gives rise to challenges when one wants to process the resulting non-rectangular pixel grids with convolutions. \cite{Cohen17} compute convolutions on spherical harmonics, but such frequency-domain networks do not yet reach the accuracy of regular CNNs. 
A common workaround is to unfold the spherical images along the azimuth and longitude dimensions, which leads to lesser artifacts than perspective projection to a planar image. Nevertheless, extreme stretching at the sphere poles remains. 
This deformation has been handled by learning filters that have the same response as processing local planar patches~\cite{Su17} or by using Deformable Convolutional Networks~\cite{Dai17}.
However, any convolution is location invariant and misses the geometric position that caused the deformation. 
To counteract this,~\cite{Liu18} propose to add the pixel coordinates as an input feature. In our preliminary experiments, however, we observed this to lead to overfitting and degraded results.

Convolution can be defined directly on the sphere, by sampling points reflecting the sphere curvature \cite{Khasanova17,Coors18}. This leads to high accuracy and position invariance, but is computationally expensive because
non-regular convolution kernels are needed at each neural network layer, which hampers parallelization and cache efficiency.
By contrast, we target a single undistortion layer that works in harmony with regular CNNs, MLPs, and attention mechanisms.

\paragraph{Attention windows.} Processing RoIs instead of the entire input image leads to computationally more efficient and more accurate models. Most prominent and related to our approach are Spatial Transformer Networks (STN) \cite{Jaderberg15} that learn invariance to translation, scale, rotation and more generic warping by spatially transforming the feature maps with an affine or free-form deformation. Multiple STNs have also been stacked \cite{Lin17e} to model more complex transformations.
STNs proceed in two steps: First, a grid of sample points is defined in the original image, either by direct regression or by predicting the parameters of a restricted family of transformations, such as a $3\times3$ matrix for affine transformations.
Second, the pixel value at each grid point is mapped to the target by bilinear interpolation of the neighboring image pixels. This yields differentiability and enables end-to-end training as an ordinary layer within deep network architectures. It also applies to 3D transformations \cite{Yan16}. In this work, we generate a sampling grid that undoes perspective effects in the RoI and use the STN to maintain differentiability with respect to the RoI position and scale.

\section{Perspective Crop Layer}
We start our derivation by formulating local processing and existing cropping solutions mathematically as an affine transformation between a real and virtual camera of fixed orientation. Subsequently, we derive the perspective transformation underlying PCL, which corresponds to a rotation of the virtual camera frame, and finally introduce the implementation of PCL via two neural network layers that sandwich the backbone prediction network. 

\subsection{Motivation and Rectangular Crops} 

\label{sec:motivation}

Each point $\hat{\vq}$ of a rescaled rectangular or trapezoidal image patch can be expressed in terms of the original image coordinates $\vq$. This affine transformation can be written in projective coordinates as
\begin{equation}
\hat{\vq} = \mC \vq \;,\;\mbox { with } \mC =
\begin{bmatrix}
s_x & c_x & a_x \\
c_y & s_y & a_y \\
0 & 0 & 1
\end{bmatrix} \; , 
\label{eq:affine}
\end{equation}
where $\va=[a_x,a_y]$ defines a 2D translation,  $\vs=[\vs_x,\vs_y]$ are scalings in two different directions, and $\vc=[c_x,c_y]$ are skew parameters. 
Therefore, as shown in Fig.~\ref{fig:virtual camera}, a cropped image can be thought of as being taken by a virtual camera with intrinsic parameters $\Kvirt = \mC \mK$, where $\mK$ is the true  $3\times3$ matrix of intrinsic parameters. 
As the translation $\va$ is usually chosen so that the patch contains an object of interest, the optical center of the virtual camera depends on the target location, which means that objects projected far from the image center are deformed differently from those near it, as shown in Fig.~\ref{fig:teaser}. To remedy this, our goal is to design a crop operation such that the optical center of the  virtual camera is always at the center of the patch, which makes perspective distortion independent from image location.  

The centering of 2D human pose commonly done in the state-of-the-art 2D-to-3D lifting approaches is a form of rectangular cropping, too. A pose is root-centered by multiplication with an affine matrix $\mC$, in which  $\va$ is the pelvis/root position, $\vs = 1$ and $\vc = 0$.
The subsequent root-centered processing with an MLP has the same downsides as cropping in STNs and convolution in CNNs in that information about the image location is removed while being affected by position-dependent perspective effects.

\subsection{Defining a Virtual Camera}

\begin{figure}
 	\centering
	\includegraphics[width=1\linewidth,trim={0 0 0 0},clip]{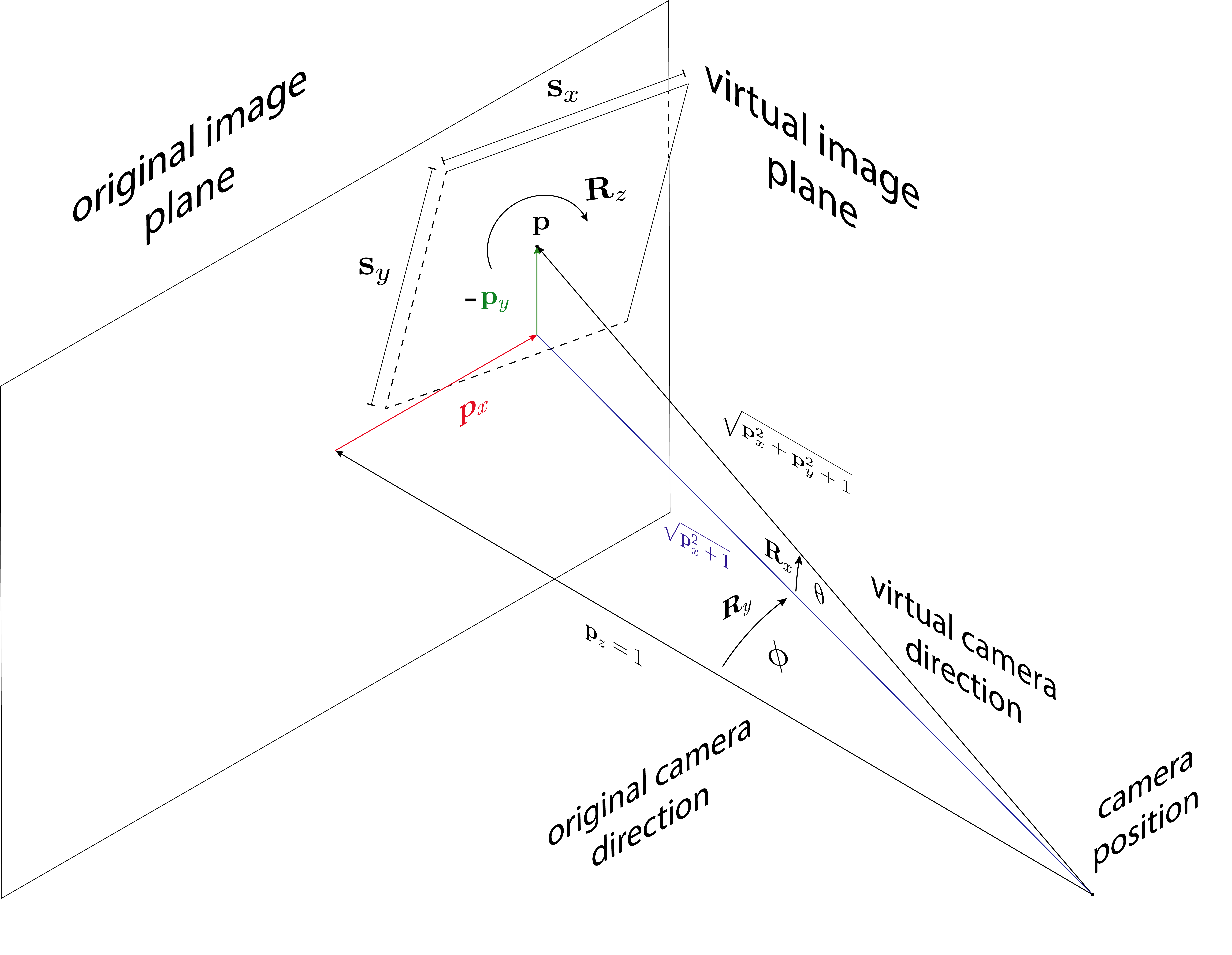}
	\caption{\label{fig:virtual camera}%
		{\bf Virtual camera.} Underlying to PCL is the projection from the original image plane onto a virtual camera pointing at the crop location. This figure visualizes the various quantities needed to infer the mapping.
	}
\end{figure}

We introduce a virtual camera with the same optical center as the real one but whose optical axis points at the center of the target patch $\vp=[\vp_x,\vp_y]$ and whose focal length is chosen to zoom onto the region of interest with factor $\vs = [\vs_x,\vs_y]$, as shown in Fig.~\ref{fig:virtual camera}. Below, we derive the virtual extrinsic parameters, in the form of rotation matrix $\mRvr$, and virtual intrinsic parameter matrix, $\mK^\text{virt}$, such that these constraints are fulfilled.

\paragraph{Cropping as a change of camera perspective.}

We aim to find camera parameters such that mapping a pixel from the original image to the cropped patch can be done by multiplying an image coordinate in homogeneous coordinates, $(u,v,1)^{\top}$, by the matrix
\begin{equation}
\mGamma(u,v,\vs,\mK) = \mK^\text{virt} \mRrv \mK^{-1} \;.
\label{eq:warp}
\end{equation}
This transformation undoes the original projection using $\mK^{-1}$, rotates the resulting point to the virtual camera with $\mRrv$ and projects it using $\mK^\text{virt}$.
As for the typical rectangular cropping defined in Eq.~\ref{eq:affine}, mapping from the image to the patch remains a warp and does not depend on the generally unknown scene geometry. By contrast to rectangular cropping, this warp is non-linear.

\paragraph{Extrinsic Parameters.} 
Let $\mRvr$ be the $3\times3$ rotation matrix that defines the virtual camera orientation. It can be written as $\mR_{y} \mR_{x} \mR_z$,  where $\mR_x$,  $\mR_y$, and $\mR_z$ are the Euler rotation matrices that rotate counter-clockwise around the $x$, $y$, and $z$ axes of the original camera coordinate system,
as depicted by Fig.~\ref{fig:virtual camera}. 
Two degrees of freedom of $\mRvr$, $\mR_x$ and $\mR_y$, are determined by pinning the center of the virtual camera to the backprojected point $\vp = \mK^{-1} (u,v,1)^\top$ in the real camera. Formally, we compute
\begin{equation}
\small
\mRvr %
= \hspace{-0.05cm}\left[ {\begin{array}{ccc}
	\frac{1}{\sqrt{1+{\vp}_x^2}} & 
	\frac{-{\vp}_x {\vp}_y}{\sqrt{(1+{\vp}_x^2+{\vp}_y^2) (1+{\vp}_x^2)}} &
	\frac{{\vp}_x}{\sqrt{1+{\vp}_x^2+{\vp}_y^2}} \\
	0 & 
	\frac{\sqrt{1+{\vp}_x^2}}{\sqrt{1+{\vp}_x^2+{\vp}_y^2}} &
	\frac{{\vp}_y}{\sqrt{1+{\vp}_x^2+{\vp}_y^2}} \\
	\frac{-{\vp}_x}{\sqrt{1+{\vp}_x^2}} &
	\frac{-{\vp}_y}{\sqrt{(1+{\vp}_x^2+{\vp}_y^2) (1+{\vp}_x^2)}} &
	\frac{1}{\sqrt{1+{\vp}_x^2+{\vp}_y^2}}
	\end{array} } \right]\hspace{-0.05cm} \;,
\label{eq:viewRotation}
\end{equation}
The details are provided in the \supp.

The yaw angle around the optical axis is unconstrained. We set it to zero (pointing upwards) in our experiments. Instead, $\mR_{z}$ could be controlled, to normalize subject orientation to pose human subjects upright in the virtual view.

\begin{figure}[t]
	\centering
	\includegraphics[width=0.9\linewidth,trim={0 0 0 0},clip]{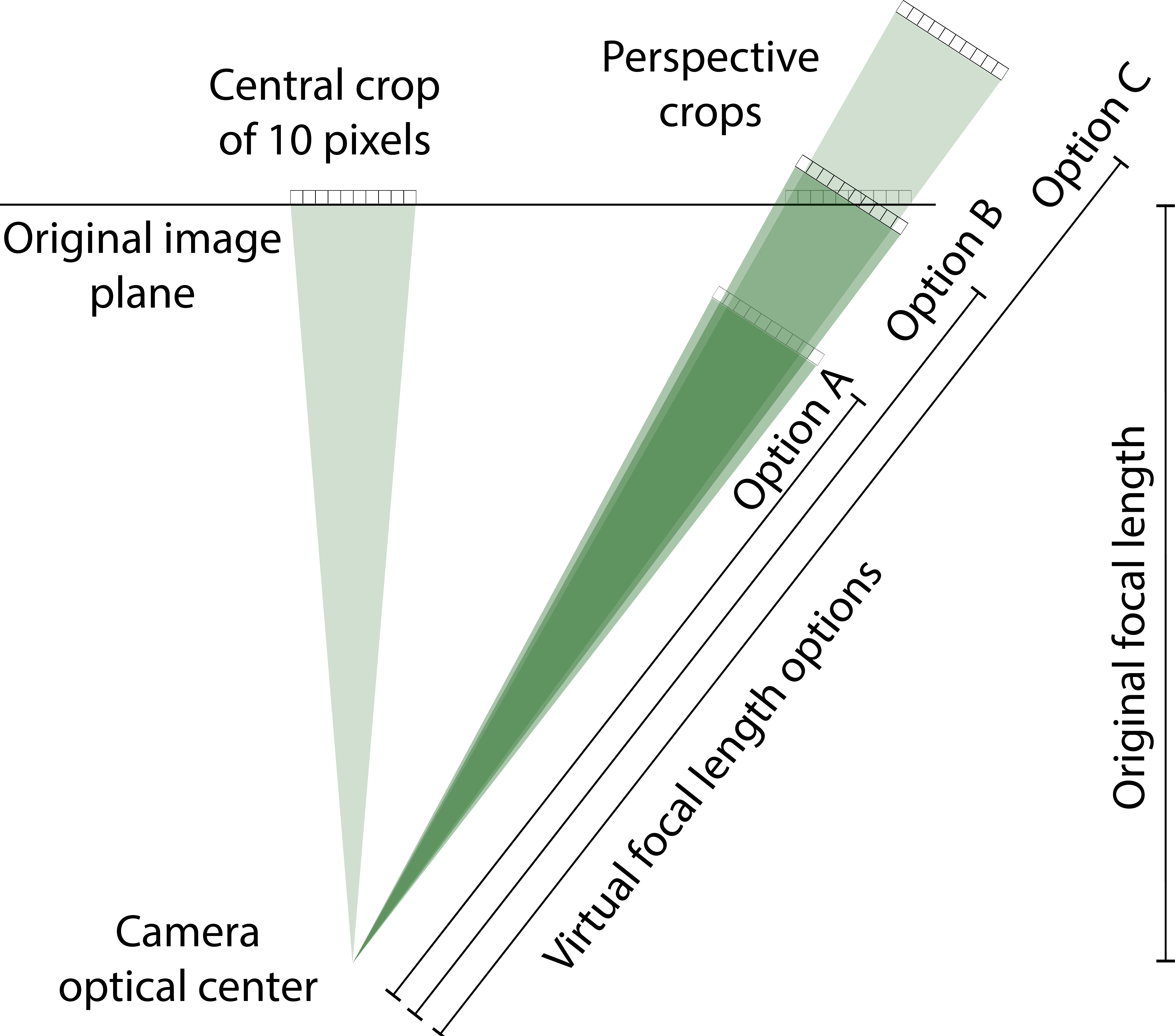}%
	\captionof{figure}{
		{\bf Focal length settings.} We propose three different variants for inferring the focal length of the camera to match the pixel scales in the original and transformed view. The intersection of the green cones with the image plane is the fraction of pixels that is cropped by PCL. Only Option C maintains a consistent scaling between center and off-center positions.}
	\label{fig:camera intrinsics}%
\end{figure}

\paragraph{Intrinsic Parameters.} 

Let
\begin{equation}
\Kvirt =
\begin{bmatrix}
\vf^\text{virt}_x & 0 & t^\text{virt}_x \\
0 & \vf^\text{virt}_y & t^\text{virt}_y \\
0 & 0 & 1
\end{bmatrix} 
\end{equation}
be the $3\times3$ matrix of intrinsic parameters of the virtual camera. Putting the optical center in the middle of the patch means that $t^\text{virt}_x =t^\text{virt}_y = 0.5$. The virtual focal lengths are 
$\vf^\text{virt} = [\vf_x^\text{virt} , \vf_y^\text{virt} ]= \frac{\vh^\text{virt}}{\vs}$, where $\vh^\text{virt}$ is a function of the original focal length in the horizontal and vertical direction stored in $\mK$, and $\vs$ determines the crop scale in relation to the full image. Together with $\vp$, it defines the area of interest and is the input to PCL. 

There is no universal way for choosing $\mathbf{\vh^\text{virt}}$, the virtual camera's focal length, without scaling to the smaller crop size. We propose the following three alternatives and evaluate their influence empirically in Section \ref{sec:Experiments}:

\begin{itemize}
	\item[\textbf{A.}] Setting $\mathbf{\vh^\text{virt}}$ to $\mathbf{f}$, the original focal length.
    \item[\textbf{B.}] Setting $\mathbf{\vh^\text{virt}}$ to $\mathbf{f} \|{\vp}\|$, so that the virtual image plane intersects with the real one at $\vp$.
	\item[\textbf{C.}] Setting $\mathbf{\vh}^\text{virt}_x = \vf_x  \|{\vp}\| \sqrt{\vp_x^2+1}$ and $\vh^\text{virt}_y = \vf_y \frac{\|{\vp}\|^2}{\sqrt{\vp_x^2+1}}$ to preserve pixel scales.
\end{itemize}

\begin{figure}[t]
	\centering
	\includegraphics[width=1\linewidth]{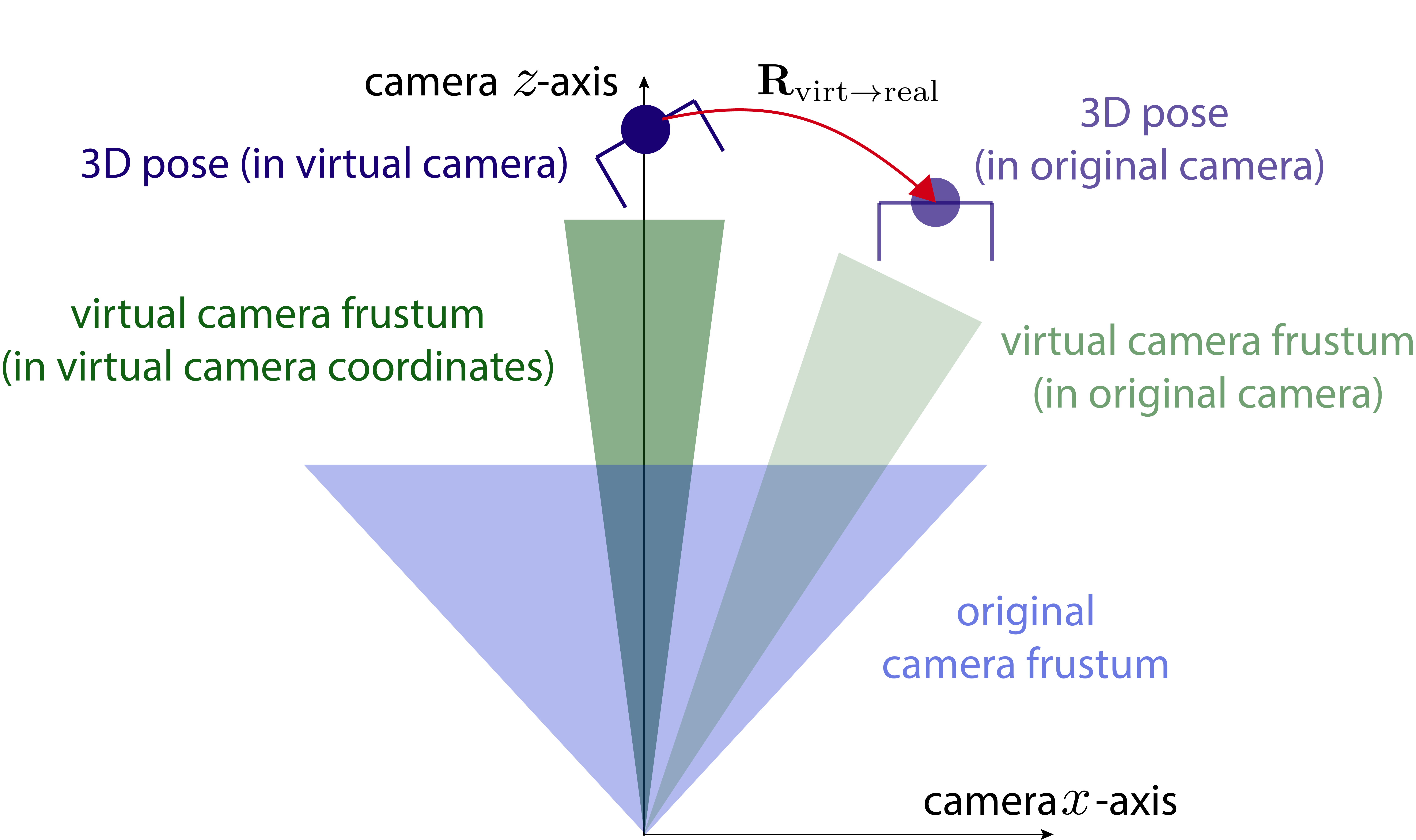}%
	\captionof{figure}{{\bf 3D Pose change.} Because networks equipped with a PCL layer operate in a virtual camera, the 3D pose prediction living in these coordinates is transformed back from virtual to original camera coordinates in the PCL\_inv layer by applying $\mRvr$.}
	\label{fig:pose_change}%
\end{figure}

Although the nature of such a perspective transformation cannot maintain scale in all parts of the image, the last choice of parameters guarantees that scale between the original and virtual image is preserved along the vertical and horizontal axis, while scaling non-linearly in the diagonal direction. Fig.~\ref{fig:teaser} visualizes this behavior, and Fig.~\ref{fig:camera intrinsics} illustrates the crop width after with each of the three choices.

\subsection{Differentiable Network Layer}
\label{sec:pcl implementation}

\begin{figure}
\begin{center}
	\includegraphics[width=1\linewidth]{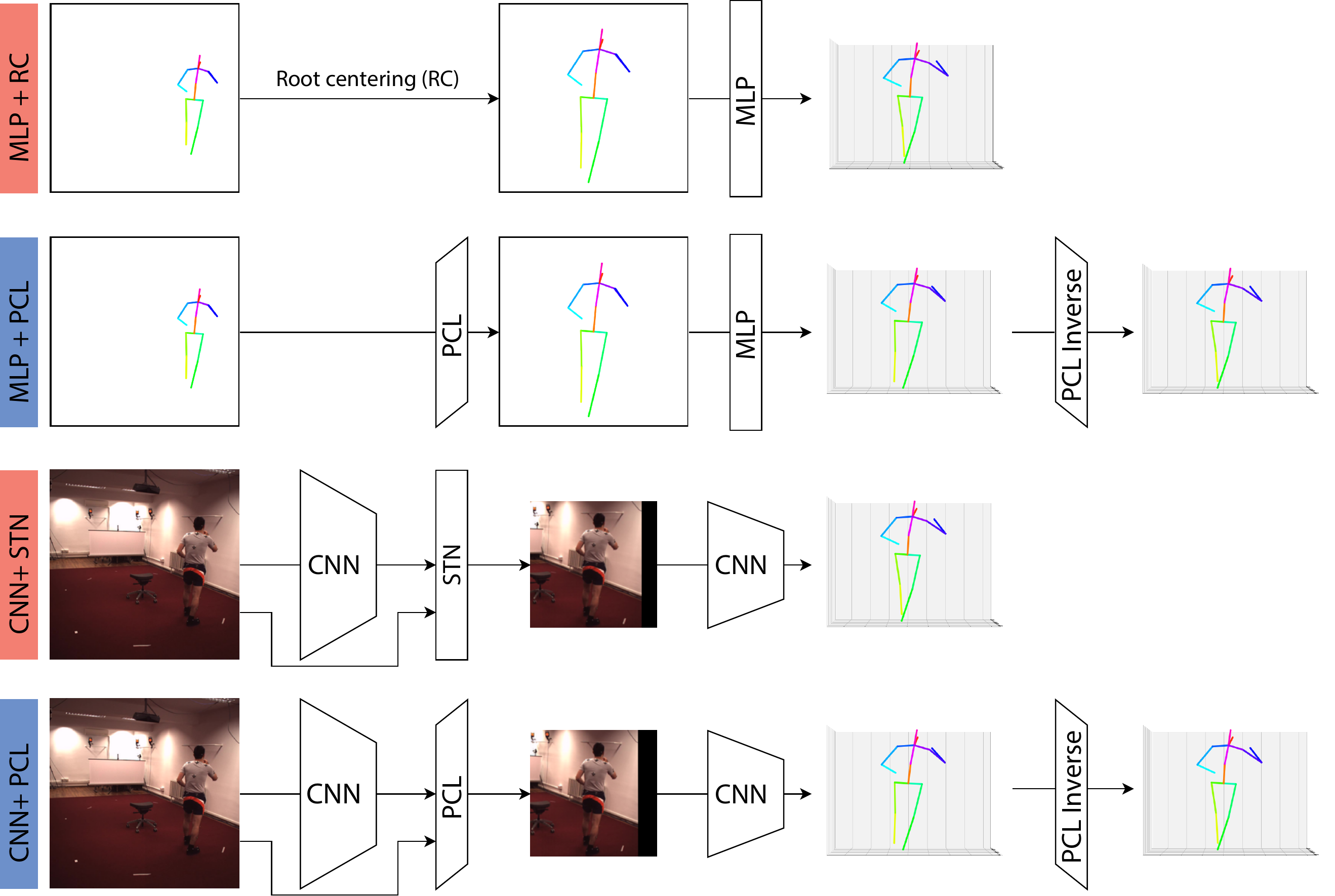}%

	\caption{\label{fig:network architectures}%
		{\bf PCL integration into existing architectures.} PCLs are applied in pairs, sandwiching the original neural network backbone with the PCL mapping from original to virtual image space and the PCL\_inv that maps 3D predictions back to the original camera coordinates. 
	}
\end{center}
\end{figure}

PCLs are designed to facilitate perspective correction within existing deep neural network architectures. We propose the two forms that are depicted in Fig.~\ref{fig:network architectures}. Two layers are involved, the projection to the virtual camera and the transformation of the reconstruction to the original camera.

\paragraph{PCL for lifting 2D keypoints to 3D with MLPs.} For networks taking 2D keypoints as input, such as the locations of the human body parts detected in the image plane, Eq.~\ref{eq:warp} can be applied directly on every 2D coordinate and becomes a simple pre-processing that normalizes the 2D pose for perspective effects. The target center location $\vp$ can be chosen as the mean of all joints, or a root joint. We use the pelvis location as crop target for human pose estimation.

\paragraph{PCL for CNNs.}
Applying PCL to CNNs requires a two-stage CNN architecture. First, one or more RoIs $(\vp,\vs)^N_{i=1}$ are predicted from the input image $\mI \in \R^{W \times H \times F}$ using a detection network. Subsequently, the input is cropped to focus the attention of the subsequent reconstruction network.
PCL replaces the cropping by implementing Eq.~\ref{eq:warp} and Eq.~\ref{eq:viewRotation}. The pixels are warped using bilinear interpolation, as in the conventional STNs~\cite{Jaderberg15} we introduced in the related work section.
Because the transformation $\mGamma$, the definition of $\mRvr$, and the virtual camera matrix $\Kvirt$ rely on simple algebraic operations, the entire process is analytically differentiable. To improve efficiency and numerical stability, we parametrize $\mRvr$ in terms of length measures (Eq.~\ref{eq:viewRotation}) instead of angles and computationally-expensive trigonometric functions in the general definition of Euler angles. The derivation and relation of both are detailed in the \supp.

\paragraph{PCL\_inv: back-transformation to the real camera.} The derived
perspective crop has the regular grid structure of a normal
image or feature map. Any neural image processing steps
can be performed on it as is. However, the derived quantities
will live in the coordinates of the virtual camera. As for
classical attention windows, if the spatial context is important,
the processed crop needs to be translated back to the original camera coordinates. 
For 3D quantities, such as 3D human pose, this
amounts to applying $\mRvr$ on the reconstruction,
as shown in Fig.~\ref{fig:pose_change}. This PCL\_inv layer is the same for MLPs and CNNs.

\comment{
Let $\cT$ be a conventional spatial transformer attention window. It takes the desired 2D transformation as input and creates one sample point in the original image for each pixel in the crop. Since the sample points will in general not be aligned with the discretization of the original image, bilinear interpolation is used to map from source to target. The transformation is either affine, as in Eq.~\ref{eq:affine}, or a free-form warp field. To introduce PCL into an existing network architecture, we replace the affine sampling strategy, and generate grid points according to 
\begin{equation}
\vq = \mGamma^{-1} \hat{\vq};,
\label{eq:inverse pixel relation}
\end{equation}
where $\hat{\vq}$ is the target pixel location in the perspective crop and $\vq$ the generated source sample point.
Note that this transformation is in projective coordinates and that translation to Cartesian coordinates is a non-linear transformation. 
General deformations are still supported; they can be applied independently in the undistorted and normalized crop or be immediately chained with $\mGamma$. %
}

\comment
{
\subsection{Appendix: Virtual Camera Derivation}
\label{sec:derivation}

\paragraph{Intrinsic parameters}
Here, we assume that the image is parametrized in normalized coordinates; from 0 to 1, starting in the bottom left corner. The offset $\vt$ and focal length $\vf$ in the original intrinsic matrix, \TODO{Check if correct?}
\begin{equation}
\mK =
\begin{bmatrix}
\vf_x & \gamma & \vt_x \\
0 & \vf_y & \vt_y \\
0 & 0 & 1
\end{bmatrix}
\end{equation}
are also defined in these normalized coordinates. %
By this choice, it is possible to fix the virtual offset $\vt^\text{virt} = (0.5,0.5)$ independent of the desired crop resolution.

To maintain the scales in the original image as best as possible, the virtual focal length,
$\vh^\text{virt} = \|{\vp}\| \frac{\vf}{\vs}$,
needs to depend on the cop location. It is set so that
the zoom-level scales proportionally by $\vs$ and virtual image plane intersects with the real one in $\vp$ when $\vs=\mathbf{1}$, which ensures that pixels size in crop are in a direct relation to the ones in the original image, as visualized in 
Figure~\ref{fig:camera intrinsics}.

\paragraph{Extrinsic parameters}
The position, and therefore center of projection, of the virtual camera is not altered, as it would be impossible to re-render the orignal image in the virtual camera. Only pure rotations and zoom by a change of focal length can be modeled by warping the original image.

We seek to define the rotation matrix $\mR$ of the virtual camera so as to rotate the original point $\vp$ to be at the new image center. We construct it through Euler rotation matrices, 
\begin{align}
&\sin(\phi) = \frac{-{\vp}_x}{\sqrt{1+{\vp}_x^2}}\;,\;
\cos(\phi) = \frac{1}{\sqrt{1+{\vp}_x^2}}\;,\;
\sin(\theta) = \frac{{\vp}_y}{\sqrt{1+{\vp}_x^2+{\vp}_y^2}}\;,\;
\cos(\theta) = \frac{\sqrt{1+\vp_x^2}}{\sqrt{1+{\vp}_x^2+{\vp}_y^2}}\;,\nonumber\\
&R_y =  \left[ {\begin{array}{ccc}
	\cos(\phi) & 0 & \sin(\phi)\\
	0& 1 & 0\\
	-\sin(\phi) & 0 & \cos(\phi)\\
	\end{array} } \right], 
&R_x = \left[ {\begin{array}{ccc}1 & 0 & 0\\
	0 & \cos(\theta) & -\sin(\theta)\\
	0 & \sin(\theta) & \cos(\theta)\\
	\end{array} } \right],
\end{align}
that model rotations around the $y$- and $z$-axis by $\theta$ and $\phi$, respectively. 
These trigonometric functions of the angles depend in closed form on the target position $\vp$:
\begin{align}
\sin(\phi) &= \frac{-{\vp}_x}{\sqrt{1+{\vp}_x^2}},&
\cos(\phi) &= \frac{1}{\sqrt{1+{\vp}_x^2}},\nonumber\\
\sin(\theta) &= \frac{{\vp}_y}{\sqrt{1+{\vp}_x^2+{\vp}_y^2}},&
\cos(\theta) &= \frac{\sqrt{1+\vp_x^2}}{\sqrt{1+{\vp}_x^2+{\vp}_y^2}} \;.
\label{eq:trig}
\end{align}
Put together, Eq.~\ref{eq:euler} and Eq.~\ref{eq:trig} yield

Combination of rotation matrices
\begin{align}
&\mR_\text{virt $\to$ orig} = R_{-x} R_{-y} \\
& =
\left[ {\begin{array}{ccc}
	\cos(-\phi) & 0 &  \sin(-\phi) \\
	\sin(-\phi)\sin(-\theta) & \cos(-\theta) & -\sin(-\theta)\cos(-\phi) \\
	-\cos(-\theta)\sin(-\phi) & \sin(-\theta)&  \cos(-\phi)\cos(-\theta)\\
	\end{array} } \right] \nonumber\\
& =
\left[ {\begin{array}{ccc}
	\frac{1}{\sqrt{1+{\vp}_x^2}} & 0 &  \frac{{\vp}_x}{\sqrt{1+{\vp}_x^2}} \\
	\frac{-{\vp}_x}{\sqrt{1+{\vp}_x^2}}\frac{{\vp}_y}{\sqrt{1+{\vp}_x^2+{\vp}_y^2}} & \frac{\sqrt{1+\vp_x^2}}{\sqrt{1+{\vp}_x^2+{\vp}_y^2}} & \frac{{\vp}_y}{\sqrt{1+{\vp}_x^2+{\vp}_y^2}}\frac{1}{\sqrt{1+{\vp}_x^2}} \\
	\frac{\sqrt{1+\vp_x^2}}{\sqrt{1+{\vp}_x^2+{\vp}_y^2}}\frac{-{\vp}_x}{\sqrt{1+{\vp}_x^2}} & \frac{-{\vp}_y}{\sqrt{1+{\vp}_x^2+{\vp}_y^2}}&  \frac{1}{\sqrt{1+{\vp}_x^2}}\frac{\sqrt{1+\vp_x^2}}{\sqrt{1+{\vp}_x^2+{\vp}_y^2}}\\
	\end{array} } \right] \nonumber\\
& =
\left[ {\begin{array}{ccc}
	\frac{1}{\sqrt{1+{\vp}_x^2}} & 0 &  \frac{{\vp}_x}{\sqrt{1+{\vp}_x^2}} \\
	\frac{-{\vp}_x {\vp}_y}{\sqrt{(1+{\vp}_x^2+{\vp}_y^2) (1+{\vp}_x^2)}} & \frac{\sqrt{1+{\vp}_x^2}}{\sqrt{1+{\vp}_x^2+{\vp}_y^2}} & \frac{{\vp}_y}{\sqrt{(1+{\vp}_x^2+{\vp}_y^2) (1+{\vp}_x^2)}} \\
	\frac{-{\vp}_x}{\sqrt{1+{\vp}_x^2+{\vp}_y^2}}& \frac{-{\vp}_y}{\sqrt{1+{\vp}_x^2+{\vp}_y^2}}&  \frac{1}{\sqrt{1+{\vp}_x^2+{\vp}_y^2}}\\
	\end{array} } \right] \nonumber\\
\end{align}
Or the inverse direction (is more compact to write down)
\begin{align}
&\mR_\text{orig $\to$ virt} 
= R_y R_x \nonumber\\
& =
\left[ {\begin{array}{ccc}
	\cos(\phi) & \sin(\phi)\sin(\theta) & \cos(\theta)\sin(\phi)\\
	0 & \cos(\theta) & -\sin(\theta)\\
	-\sin(\phi) & \sin(\theta)\cos(\phi) & \cos(\phi)\cos(\theta)\\
	\end{array} } \right] \nonumber\\
& =
	\left[ {\begin{array}{ccc}
	\frac{1}{\sqrt{1+{\vp}_x^2}} & 
	\frac{-{\vp}_x {\vp}_y}{\sqrt{(1+{\vp}_x^2+{\vp}_y^2) (1+{\vp}_x^2)}} &
	 \frac{-{\vp}_x}{\sqrt{1+{\vp}_x^2+{\vp}_y^2}} \\
	0 & 
	\frac{\sqrt{1+{\vp}_x^2}}{\sqrt{1+{\vp}_x^2+{\vp}_y^2}} &
	\frac{-{\vp}_y}{\sqrt{1+{\vp}_x^2+{\vp}_y^2}} \\
	\frac{{\vp}_x}{\sqrt{1+{\vp}_x^2}} &
	\frac{{\vp}_y}{\sqrt{(1+{\vp}_x^2+{\vp}_y^2) (1+{\vp}_x^2)}} &
	\frac{1}{\sqrt{1+{\vp}_x^2+{\vp}_y^2}}
	\end{array} } \right] \;.
\end{align}

Wit x, y positive
{
	\begin{align}
	R_x R_y &= \left[ {\begin{array}{ccc}1 & 0 & 0\\
		0 & \cos(\theta) & -\sin(\theta)\\
		0 & \sin(\theta) & \cos(\theta)\\
		\end{array} } \right]
	\left[ {\begin{array}{ccc}
		\cos(\phi) & 0 & \sin(\phi)\\
		0& 1 & 0\\
		-\sin(\phi) & 0 & \cos(\phi)\\
		\end{array} } \right] \nonumber\\
	&=
	\left[ {\begin{array}{ccc}
		\cos(\phi) & 0 &  \sin(\phi) \\
		\sin(\phi)\sin(\theta) & \cos(\theta) & -\sin(\theta)\cos(\phi) \\
		-\cos(\theta)\sin(\phi) & \sin(\theta)&  \cos(\phi)\cos(\theta)\\
		\end{array} } \right] \nonumber\\
	\end{align}
}
}

\section{Experiments}
\label{sec:Experiments}

We evaluate the improvements brought about by PCL on the task of 3D human pose estimation from either images or 2D keypoints, and show that they hold for neural networks of diverse complexity. The benefits of PCL for the 2D to 3D lifting task on Human 3.6 Million dataset ~\cite{Ionescu14a} and MPI-INF-3DHP dataset ~\cite{mono-3dhp2017} are shown qualitatively in both Fig.~\ref{fig:2d3d-qualitative} and in additional experiments in the supplemental video.

\begin{figure}
\begin{center}
\includegraphics[width=1\linewidth,trim={5.25cm 5.5cm 5cm 5cm},clip]{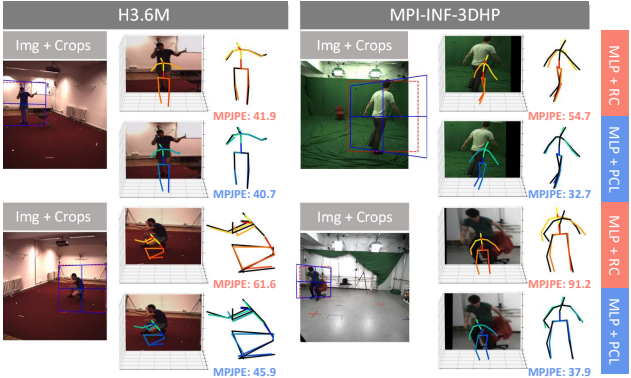}
\end{center}
\caption{\label{fig:2d3d-qualitative}
		{\bf Qualitative examples.} For both, H3.6M (left) and MPI-INF-3DHP (right), PCL improves 3D pose estimation significantly by predicting the orientation of limbs more precisely. The MLP+PCL output is shown in blue and the baseline w/o PCL in red. Individual MPJPE scores (in mm) are reported for each, which relates the visual and quantitative improvements.
	}
\end{figure}

\paragraph{Baselines.} We integrate PCL into the three neural network architectures for 3D pose estimation discussed below, and compare the resulting networks with the original ones.

\textbf{MLP+RC}: As a first baseline, we use the four-layer MLP from \cite{Martinez17} with root centering. To ensure a fair comparison, we scale the 2D input of the baseline by the crop scales $\vs$ that are used in PCL.

\textbf{T-CNN}: Our second baseline consists of the temporal convolution 2D to 3D lifting approach of Pavllo et al.~\cite{pavllo20193d}, which operates on pose sequences. To date, it is the most accurate method in its class.

\textbf{CNN+STN}: ResNet~\cite{he2016identity} is the most widely used backbone for predicting 3D pose from images. As baseline, we use an STN that takes a $265\times 265$ image as input and outputs a $128\times 128$ patch. We do almost all tests with ground-truth crop locations determined by bounding box annotations in the dataset. This guarantees that the differences in performance that we measure during evaluation are entirely due to the type of cropping and not differences in a detector network. We test a version with an an additional L2 loss on the crop location predicted by a ResNet-18 detector network, trained end-to-end with the 3D reconstruction objective. We test ResNet backbones of depth 18 and 50.

Each baseline is extended with PCLs as follows, and as
sketched in Fig.~\ref{fig:network architectures}.

\textbf{MLP+PCL}: Our method replaces the traditional hip-centering in \cite{Martinez17} with the PCL and PCL\_inv layers.

\textbf{T-CNN+PCL}: We apply PCL to \cite{pavllo20193d} by transforming the sequence of frames to the virtual camera pointing to the image coordinates in the middle of the sequence. Note that this centering is a significant difference to the original \cite{pavllo20193d}, which works with absolute 2D positions as input. For simplicity, we assume the optical center is at the image center.

\textbf{CNN+PCL}: We simply replace the rectangular STN crop with our PCL layer, as detailed in Section~\ref{sec:pcl implementation}.

\paragraph{Datasets.} 
\textbf{H3.6M}: We evaluate the effectiveness of PCL on the popular Human 3.6 Million dataset~\cite{Ionescu14a} that features eleven subjects performing 14 different actions and provides ground-truth 3D poses and camera calibration. We use the established train/validation/test split, 17-joint skeleton, and the pre-processing of~\cite{nibali20193d}. We set the rectangular and PCL crop location to the pelvis 2D joint and compute the crop scale as the width and height of a tightly-fitting bounding box. We also experiment with using the GT depth for scale estimation.
We compare variants using 2D detections from \cite{sun2019deep, xiao2018simple} and ground truth as input for 2D to 3D lifting.
When we compare to \cite{pavllo20193d}, we use their preprocessing and train/validation/test split since consecutive frames are required. For simplicity, we assume that the image size is $1000 \times 1000$, although size varies from $1000$ to $1002$.

\textbf{MPI-INF-3DHP}: 
We also evaluate our approach on the MPI-INF-3DHP dataset~\cite{mono-3dhp2017}, which, compared to H3.6M, contains more extreme poses, outdoor environments, and is shot with wide field-of-view cameras, leading to stronger perspective effects. The cameras are calibrated, and all frames are labeled with 3D pose. We use the color augmentation from \cite{nibali20193d} and the official test set and training subjects 1-8 for training, while withholding the first sequence of subject 4 and the last sequence of subject 8 for validation.
We set the rectangular and PCL crop location to the pelvis joint for 2D to 3D lifting and at the mean of the 2D poses for the image-based variants. The crop scale is computed as the width and height of a tightly fitting bounding box. 

\textbf{ToyCube}: 
We introduce a synthetic dataset containing images of a rendered cube of edge length 0.5 m. We use this toy example to ablate individual factors of variation, such as the effect of illumination and pose distribution.

\paragraph{Training setup.} The 3D pose is trained on an L2 loss using Adam \cite{Kingma15} with a learning rate of 0.001 for the 2D to 3D lifting models and 0.0005 for the image to 3D networks. The temporal convolution networks are trained using Amsgrad \cite{reddi2019convergence} with an initial learning of 0.001 and a learning rate decay factor of 0.95 applied after each epoch. We train 2D to 3D lifting methods for 200 epochs and batch size 64 and the temporal convolution networks for 80 epochs and batch size 1024 with up to 243 frames in each batch element. Lastly, image to 3D networks using a ResNet-50 backbone are trained for 40 and 150 epochs on H3.6m and MPI-INF-3DHP respectively. The ResNet-18 backbone trained on H3.6M is trained for 60 epochs.

\paragraph{Metrics.} To quantitatively evaluate 3D pose accuracy, we use the Mean Per Joint Position Error (MPJPE), computed as the average Euclidean distance of the predicted 3D joints to the ground-truth ones, where both poses are centered at the pelvis. All MPJPE results are reported in millimeters.
We also report the percentage of correct keypoints (PCK), encoding the proportion of joints whose distance to the ground truth is less than a threshold, using thresholds of 50 and 100 millimeters.

\begin{figure}
\begin{center}
\includegraphics[width=1\linewidth]{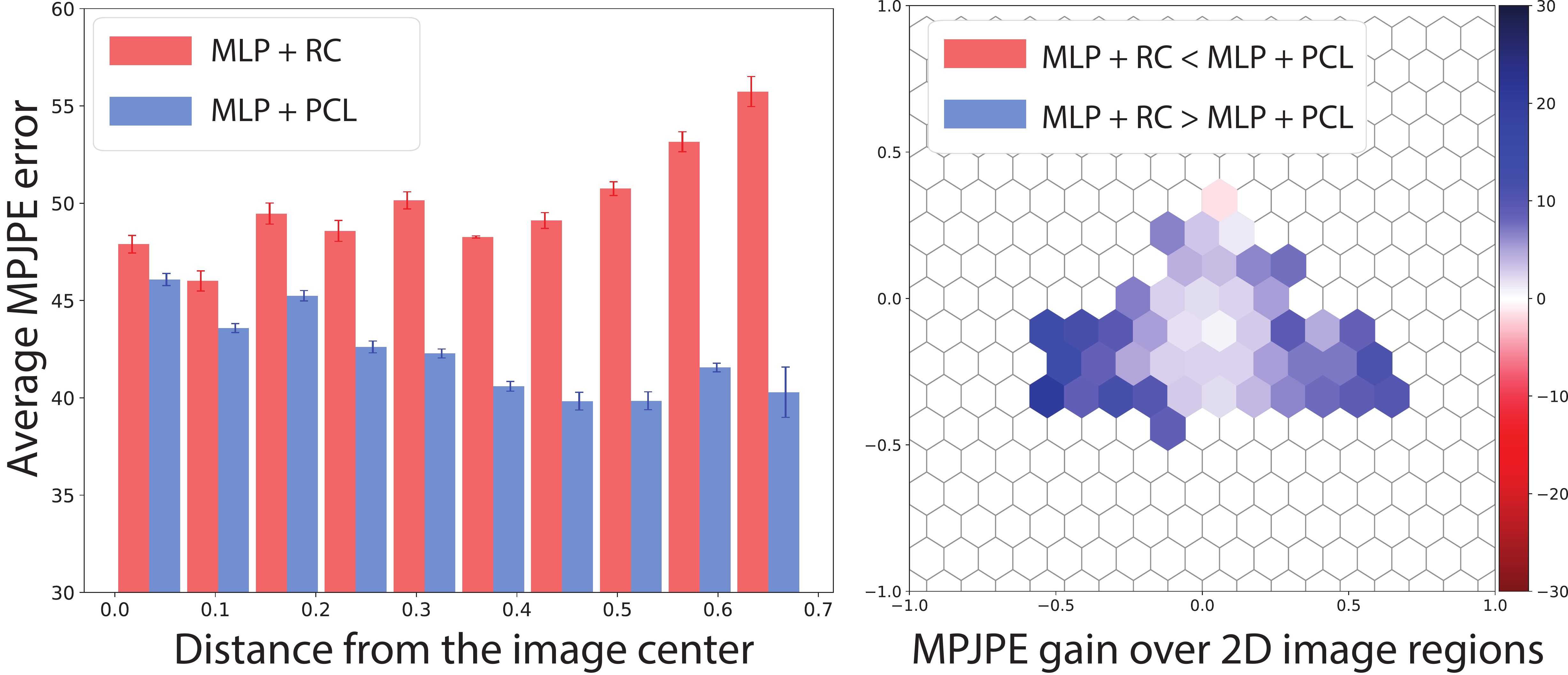}
\end{center}
\caption{\label{fig:2d3d-1d}\label{fig:2d3d-seeded-2dlocation}
		{\bf Improvement of reconstruction error,} binned with respect to the image position. \textbf{Left:} MLP+RC suffers from perspective effects away from the center, while MLP+PCL effectively compensates these leading to improvements of up to 25\%. \textbf{Right:} The consistent difference of MLP+RC and MLP+PCL is also reflected over a 2D tiling, showing the average MPJPE error difference of cells with 10 or more frames on the validation set.
	}
\end{figure}

\begin{table*}[t!]
\resizebox{\linewidth}{!}{%
\begin{tabular}{ll|ccc|ccc}
                   &                              & \multicolumn{3}{c|}{H3.6M}                                                                    & \multicolumn{3}{c}{MPI-INF-3DHP}                                                            \\
Input              & Model                        & MPJPE $\downarrow$ & PCK @ 50mm $\uparrow$ & PCK @ 100mm $\uparrow$ & MPJPE $\downarrow$ & PCK @ 50mm $\uparrow$ & PCK @ 100mm $\uparrow$ \\
\hline \hline
2D GT + 3D Root GT & MLP + RC (\cite{Martinez17})                    & $45.3 \pm 0.2$              & $66.8 \pm 0.3$                   & $91.4 \pm 0.2$                     & $69.4 \pm 0.4$              & $46.4 \pm 0.5$                   & $77.0 \pm 0.3$                    \\
2D GT + 3D Root GT & MLP + PCL (Ours)             & $\mathbf{40.1 \boldsymbol{\pm} 0.5}$     & $\mathbf{72.8 \boldsymbol{\pm} 0.5}$          & $\mathbf{93.3 \boldsymbol{\pm} 0.2}$            & $\mathbf{45.6 \boldsymbol{\pm}  0.5}$    & $\mathbf{70.1 \boldsymbol{\pm} 0.5}$          & $\mathbf{89.7 \boldsymbol{\pm} 0.3}$          \\
\hline
2D GT              & MLP + RC (\cite{Martinez17})                    & $48.4 \pm 0.4$              & $62.9 \pm 0.5$                   & $90.3 \pm 0.2$                     & $74.1 \pm 0.5$              & $43.0 \pm 0.3$                   & $74.7\pm 0.7$                     \\
2D GT              & MLP + PCL (Ours)             & $\mathbf{43.8 \boldsymbol{\pm} 0.1}$     & $\mathbf{68.3 \boldsymbol{\pm} 0.1}$          & $\mathbf{92.2 \boldsymbol{\pm} 0.0}$            & $\mathbf{50.1 \boldsymbol{\pm} 0.2}$     & $\mathbf{65.5 \boldsymbol{\pm} 0.2}$          & $\mathbf{87.8 \boldsymbol{\pm} 0.1}$           \\
\hline
2D Detection       & MLP + RC (\cite{Martinez17})                    & $69.7 \pm 0.2$              & $46.3 \pm 0.5$                   & $80.5 \pm 0.1$                     & -                         & -                              & -                               \\
2D Detection       & MLP + PCL (Ours)             & $\mathbf{67.0 \boldsymbol{\pm} 0.1}$     & $\mathbf{48.7 \boldsymbol{\pm} 0.2}$          & $\mathbf{82.1 \boldsymbol{\pm} 0.0}$            & -                & -                     & -                      \\ 
\hline
Image              & CNN (ResNet50) + STN        & 96.5                      & 32.7                           & 64.6                             & 117.7                     & 33.6                           & 60.1                            \\
Image              & CNN (ResNet50) + PCL (Ours) & \textbf{94.1}             & \textbf{34.1}                  & \textbf{65.8}                    & \textbf{109.5}            & \textbf{40.3}                  & \textbf{66.2}                   \\ \hline
Image              & CNN (ResNet18) + STN        & 95.9                      & 35.4                           & 65.6                             & -                         & -                              & -                               \\
Image              & CNN (ResNet18) + PCL (Ours) & \textbf{93.9}             & \textbf{37.1}                  & \textbf{66.6}                    & -                         & -                              & -                               \\ \hline
\end{tabular}
}
\caption{Shown are the reported MPJPE in millimeters as well as the PCK for 2D to 3D keypoint lifting tests performed on H3.6M. The reported mean and standard deviation is computed over three runs with varying random seed. For MPJPE, lower values are better and for PCK, higher values are better. We can see from the table that our method significantly outperforms the baselines that do not use PCL. We bold the best performing models in each category.}
\label{table:Quantitative_Results}
\end{table*}

\subsection{PCL for 2D to 3D Keypoint Lifting}
The results for the 2D to 3D lifting task on H3.6M and MPI-INF-3DHP are provided in Table~\ref{table:Quantitative_Results}.
For H3.6M, MLP+PCL achieves an MPJPE of 67.0 mm vs. 69.8 mm MPJPE of the MLP+RC baseline \cite{Martinez17} when using 2D detections from \cite{sun2019deep, xiao2018simple} as input, a 4\% improvement.
Even larger improvements are achieved when using the GT 2D pose as input and when using the 3D root joint position for scale estimation. Notably, our method with a scale computed from the 2D pose still outperforms the STN baseline using 3D ground truth for scale prediction.

We obtain even larger improvements with PCL on the MPI-INF-3DHP dataset, with 2.4 cm in MPJPE and 22 PCK points. This dramatic improvement is no surprise since the larger field of view (smaller focal length $f$) of the MPI-INF-3DHP cameras leads to stronger perspective effects and, therefore, a larger difference between the corrected and uncorrected views. These experiments also show that PCL is not specific to any particular focal length.

In Figure~\ref{fig:2d3d-1d}, we analyze the position dependent effect of our method on H3.6M. As shown by the plot, the baseline MPJPE increases with the distance from the subject to the image center, hinting at the negative effect of perspective distortion.
PCLs undo this effect, leading to a more stable MPJPE and outperforming the baseline by a growing margin as the distance increases. PCL even decreases with the distance to the center, which is surprising. We believe this is because the most complex poses in H3.6M, such as sitting and lying on the ground, are performed in the image center while walking dominates for the off-center ones.

\subsection{PCL for Temporal CNNs}

As shown in Table~\ref{table:Temporal_Results} incorporating PCL into the temporal convolutional network of~\cite{pavllo20193d} does not improve results on their original implementation.
Our following analysis shows that this is due to \cite{pavllo20193d} already learning position-dependent effects by operating on unnormalized 2D pose. This, however, overfits to the camera used at training time. By contrast, PCL generalizes perfectly when the camera changes at test time so long as its properties are known approximately.

\begin{table}[]
\resizebox{\linewidth}{!}{%
\begin{tabular}{l|l|ccc}
Model                   & Original $f$     & New $f$             \\ \hline \hline
T-CNN                   & $\mathbf{47.3 \pm 0.0}$ & $72.7 \pm 0.5$         \\
T-CNN + RC              & $51.5 \pm 0.1$        & $51.5 \pm 0.1$          \\
T-CNN + PCL (known $f$)    & $48.8 \pm 0.3$        & $\mathbf{48.9 \pm 0.2}$ \\
\hline
\end{tabular}
}
\caption{\textbf{Temporal CNN tests}, computed as the MPJPE over two runs with varying seed on H3.6M. While the baseline performs the best using the original camera, it is unable to generalize to new camera settings. The PCL equipped version strikes the best compromise.}
\label{table:Temporal_Results}
\end{table}

To analyze the effect, we artificially change the focal length of the test sequences by multiplying all 2D testing poses by a factor of 2/3. This simulates a camera with a smaller focal length and larger field of view. T-CNN generalizes poorly, as it seemingly overfits to the global position and scale of the training set, while PCL can adapt perfectly to the new capture setup without loss in performance.

To facilitate a fair comparison, we created a second baseline, T-CNN+RC. It centers and scales the input pose sequence by subtracting the root joint of the central frame and scaling by its horizontal and vertical size; this is the same procedure that is used for the other RC baselines. This maintains the root motion while removing absolute scale and the position in the image.
This variant generalizes much better than T-CNN does but has an overall higher error than T-CNN+PCL. In summary, PCL strikes the best compromise between accuracy on the original domain while being applicable to new camera geometries and capture setups.

\subsection{PCL for CNN Architectures} 

As shown in the last four rows of Table~\ref{table:Quantitative_Results}, when using images as input to a ResNet regressing 3D pose on H3.6M, the baselines achieves an MPJPE of 96.5 mm, while our model with PCL yields 94.1 mm, a 2.5\% reduction. 
The improvement is lower compared to 2D to 3D lifting, likely because the overall higher error for image to 3D pose prediction compared to 2D to 3D lifting is dominated by other error factors.

On the MPI-INF-3DHP dataset, PCLs' improvement is more pronounced, improving by 8mm and 6 PCK points, which further validates the previous findings that perspective effects are stronger on MPI-INF-3DHP, therefore leading to a clear improvement despite higher total errors.

\begin{figure}
\begin{center}
\includegraphics[width=0.9\linewidth]{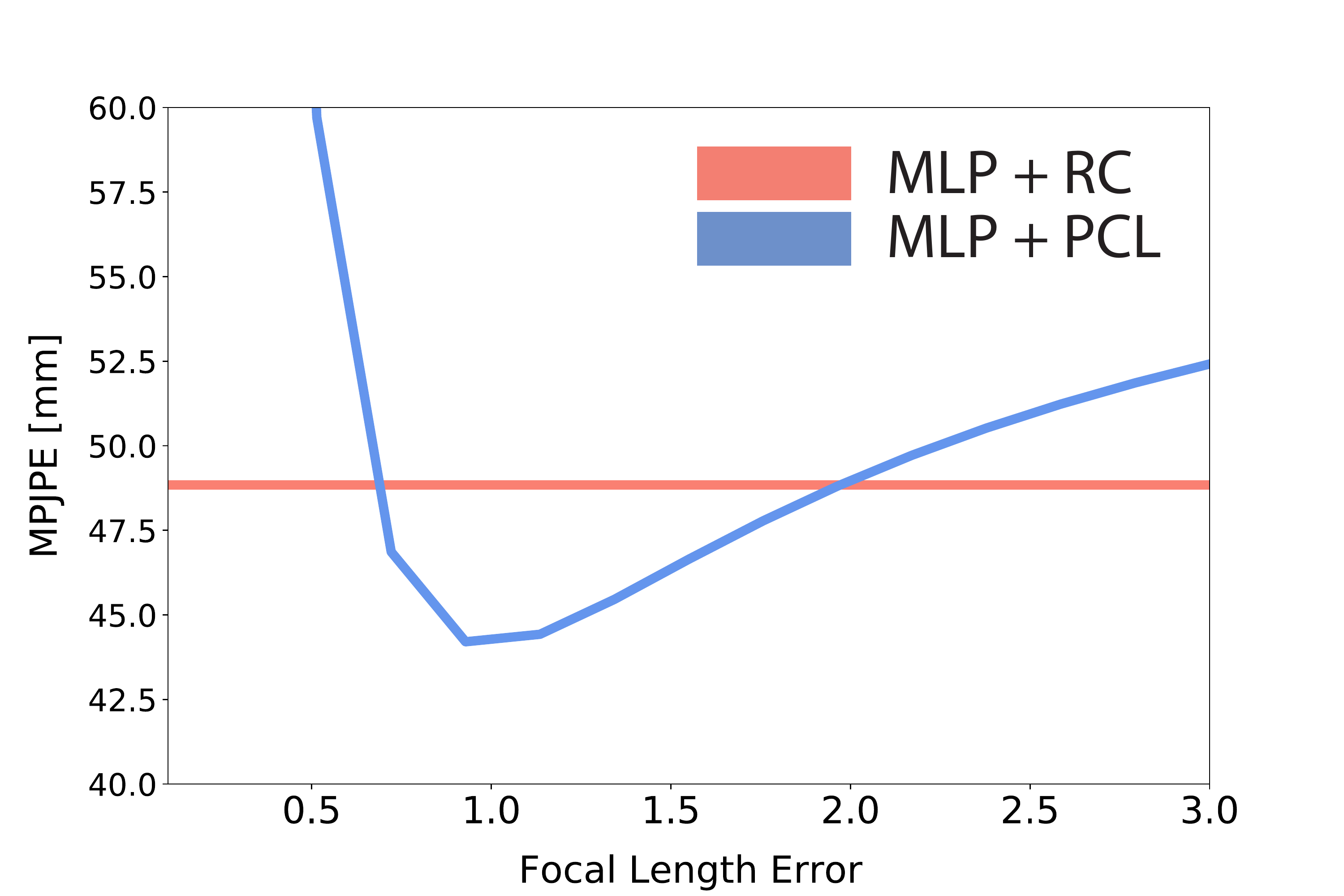}%
\end{center}
\caption{\label{fig:focal}
		{\bf Evaluating the effectiveness of PCL when the focal length of the camera is approximated.} We can see from the figure that for reasonable approximations of the focal length, PCL still outperforms the rectangular cropping. 
	}
\end{figure}

\subsection{Ablation Studies}

\textbf{Robustness to errors in focal length.} Although PCL requires knowledge of the camera's focal length, it is often known or can be estimated approximately from image features \cite{Workman15,Hold18}. To evaluate the robustness of PCL to erroneous estimates, we experiment with an artificial disturbance to the true focal length at test time. The 2D input poses to the MLP+PCL network are deformed as if they would stem from a camera with different zoom. As shown in Fig.~\ref{fig:focal}, PCL is relatively robust when the estimated focal length ranges between 0.7 and 1.5 times the true one.

\textbf{Network capacity, post-processing, and generalization.}
In addition to the main results, we i) show that the improvement by PCL is as prominent as doubling the neural network capacity from two million to four million parameters; ii) show that it is important to apply PCL at training time instead of on pre-trained models; and iii) analyze the generalizability of PCL to new unseen positions within the camera frame on a synthetic cube dataset.
Details for all these tests can be found in the supplemental document.

\section{Limitations}

The gained improvements come at the cost of requiring an estimate of the focal length $f$. Yet, the robustness towards errors in $f$ shows that improvements can still be obtained with a rough guess. It is worth to note that PCL compensates for location-dependent perspective effects. Those effects that originate from varying distance of the object to the camera, e.g., selfie vs.~third person picture, can not be resolved with image warps but would require knowledge of the 3D geometry. Data-driven approaches have been proposed to compensate these \cite{recasens2018learning,zhao2019learning}.

\section{Conclusion}

We have presented a drop-in replacement for rectangular cropping and root centering that removes location-dependent perspective effects.
It is fully differentiable, lends itself to end-to-end training, is efficient, does not impose additional network parameters, and the empirical evaluation demonstrates significant improvements for 3D pose estimation. 
Notably, the strong influence of perspective effects on the reconstruction accuracy is widely overlooked in the 3D pose reconstruction literature and these improvements are observed irrespective of the network architecture. PCL is therefore an important contribution to pushing state-of-the-art 3D reconstruction methods further.

\paragraph{Acknowledgements.}
We would like to thank Dushyant Mehta and Jim Little for valuable discussion and feedback.
This work was funded in part by Compute Canada and a Microsoft Swiss JRC project. Frank Yu was supported by NSERC-CGSM.

\appendix
\section*{Appendix}
\section{MPI-INF-3DHP Additions}

\begin{figure}
\begin{center}

  \subfloat{\includegraphics[width=0.45\linewidth]{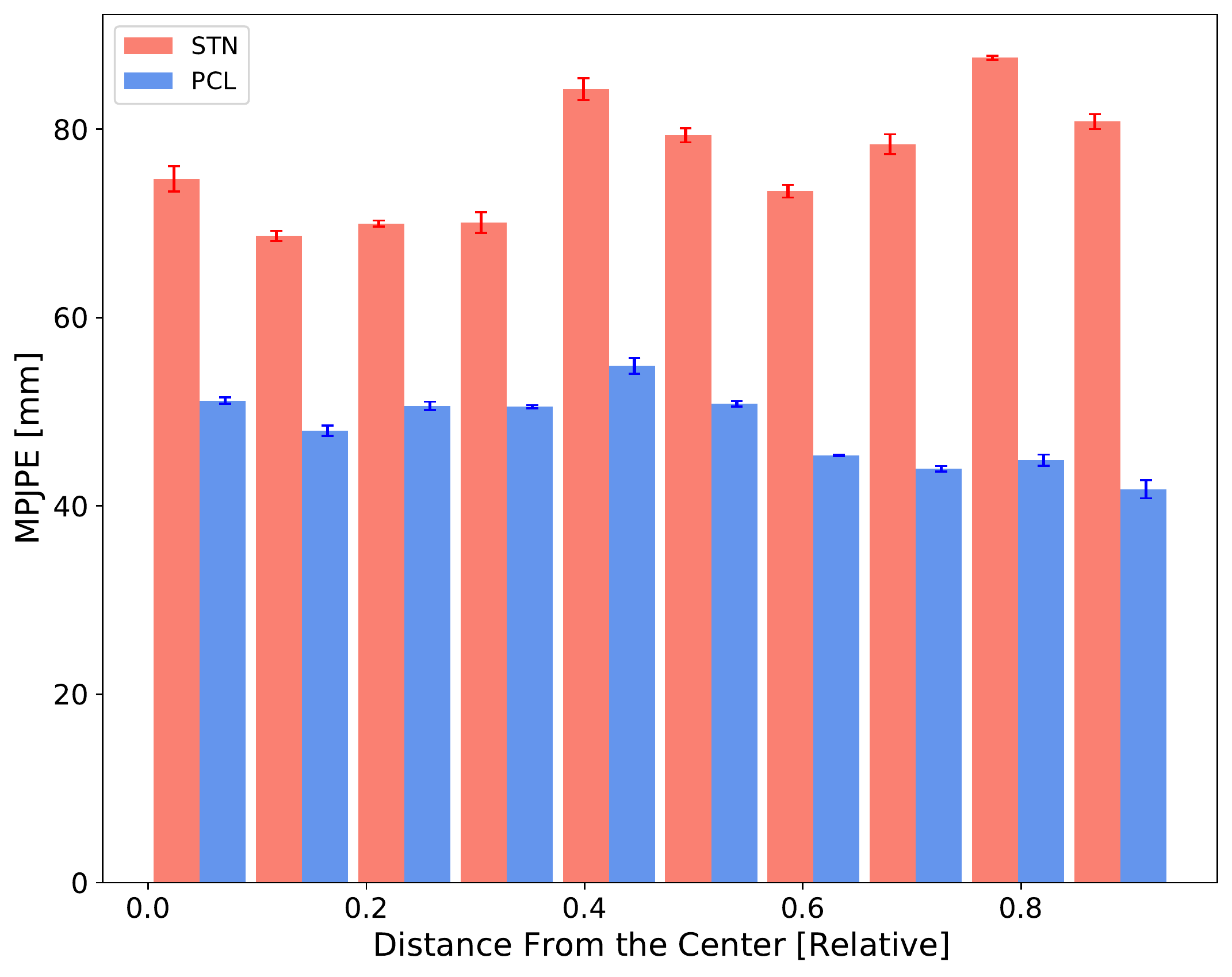}}
  \hspace{1em}%
  \subfloat{\includegraphics[width=0.45\linewidth]{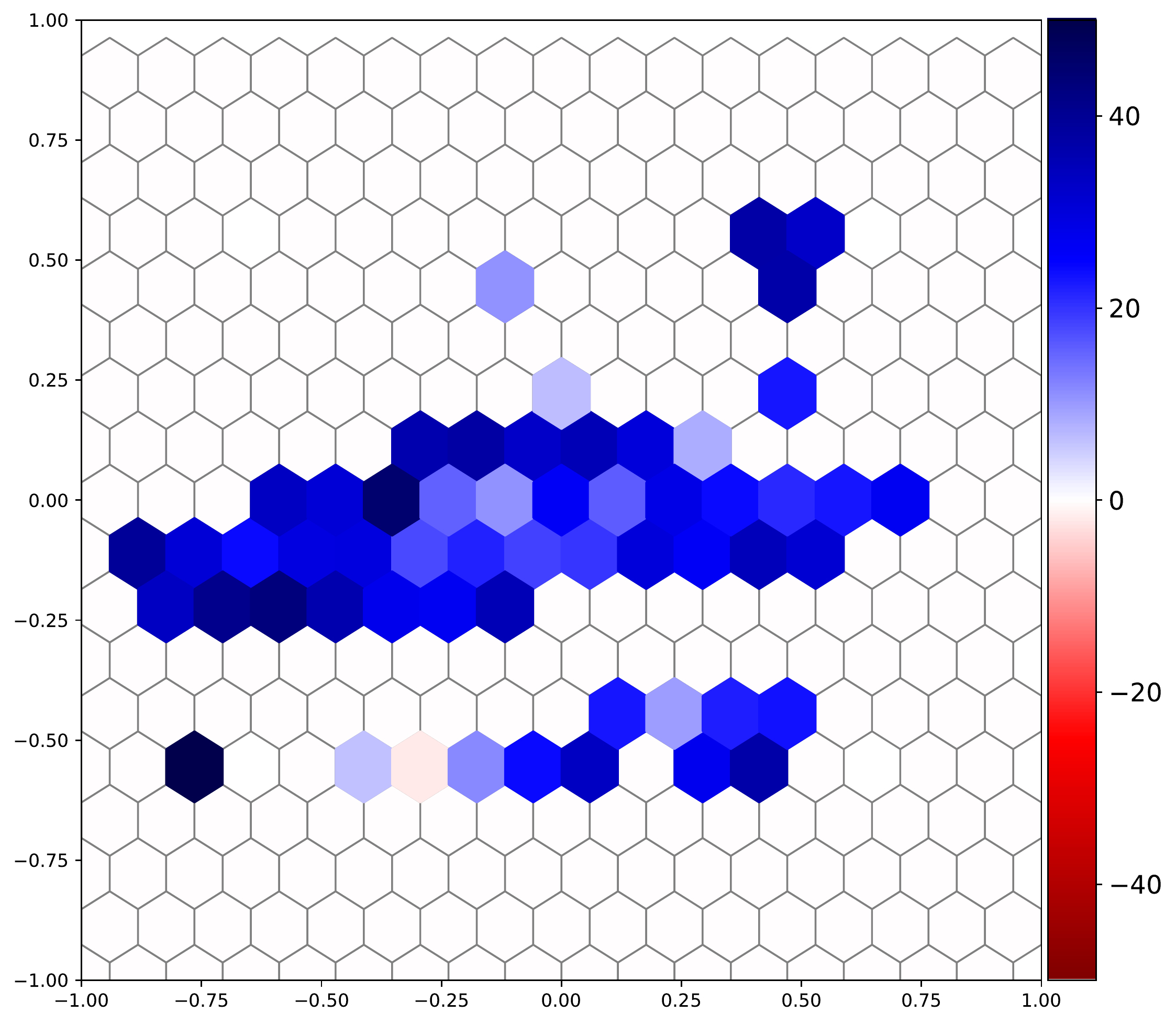}}
  \caption{\label{fig:mpi-2d3d-1d}\label{fig:mpi-2d3d-seeded-2dlocation}
		{\bf Improvement of reconstruction error,} binned with respect to the image position. \textbf{Left:} MLP+RC suffers from perspective effects away from the center, while MLP+PCL effectively compensates these leading to improvements of approximately 50\% . \textbf{Right:} The consistent difference of MLP+RC and MLP+PCL is also reflected over a 2D tiling, showing the average MPJPE error difference of cells with 10 or more frames on the validation set.
	}
\end{center}
\end{figure}

In this section, we repeat the detailed error distribution analysis done on the H3.6M dataset for the MPI-INF-3DHP dataset. Figure~\ref{fig:mpi-2d3d-1d} depicts significant improvements from using PCL. The error by PCL is stable or even improving with the distance to the center, while the MLP+RC model degrades due to perspective effects. PCL even gains an improvement at the image center. This is unexpected on the first glance but can be explained with the MPI-INF-3DHP dataset using different cameras for training and testing. The MLP+RC model seemingly overfits to the perspectives seen during training while the automatic correction of PCL leads to better generalization irrespective of the image location.

\begin{figure}
\begin{center}
  \subfloat{\includegraphics[width=0.45\linewidth]{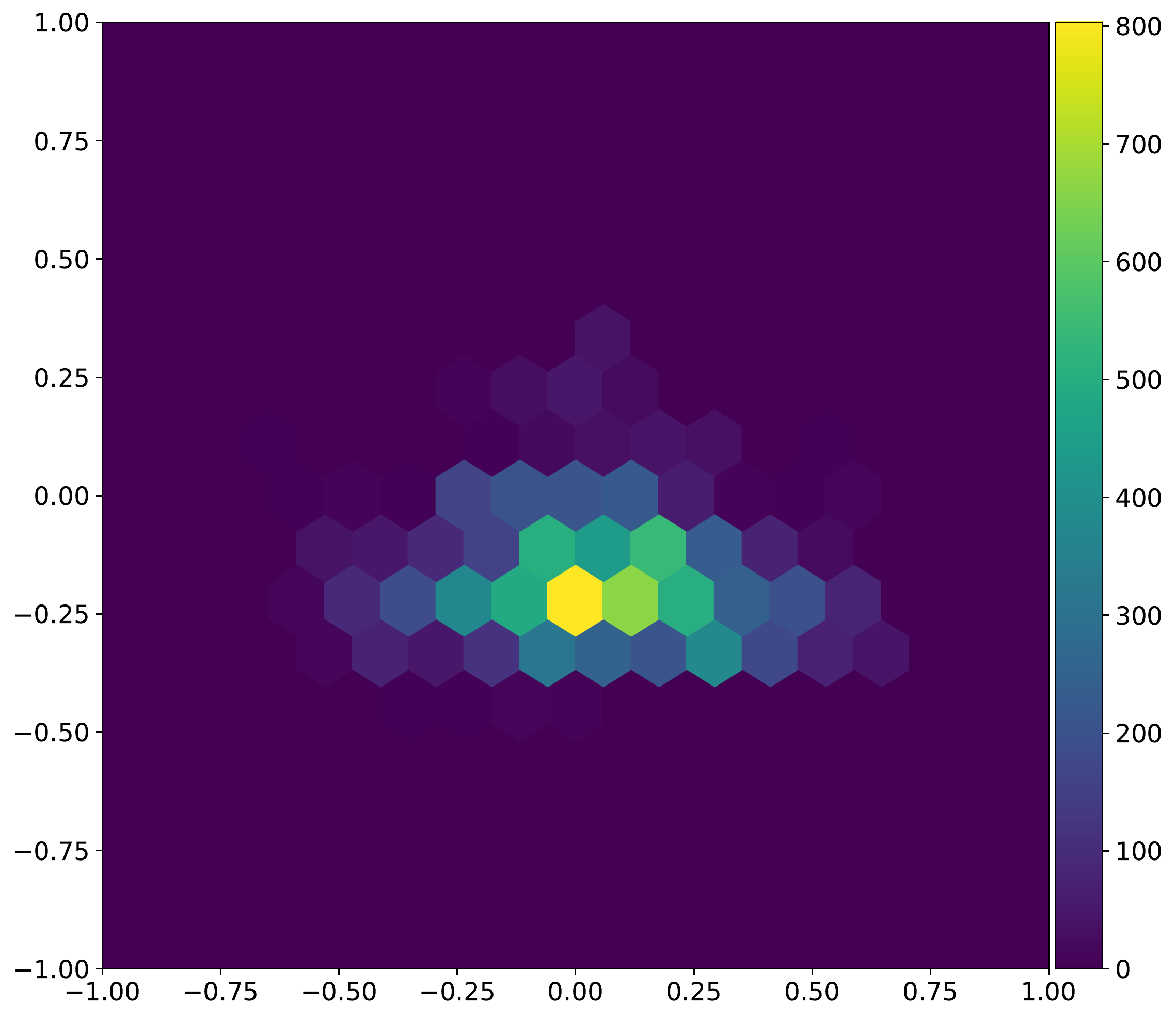}}
  \hspace{1em}%
  \subfloat{\includegraphics[width=0.45\linewidth]{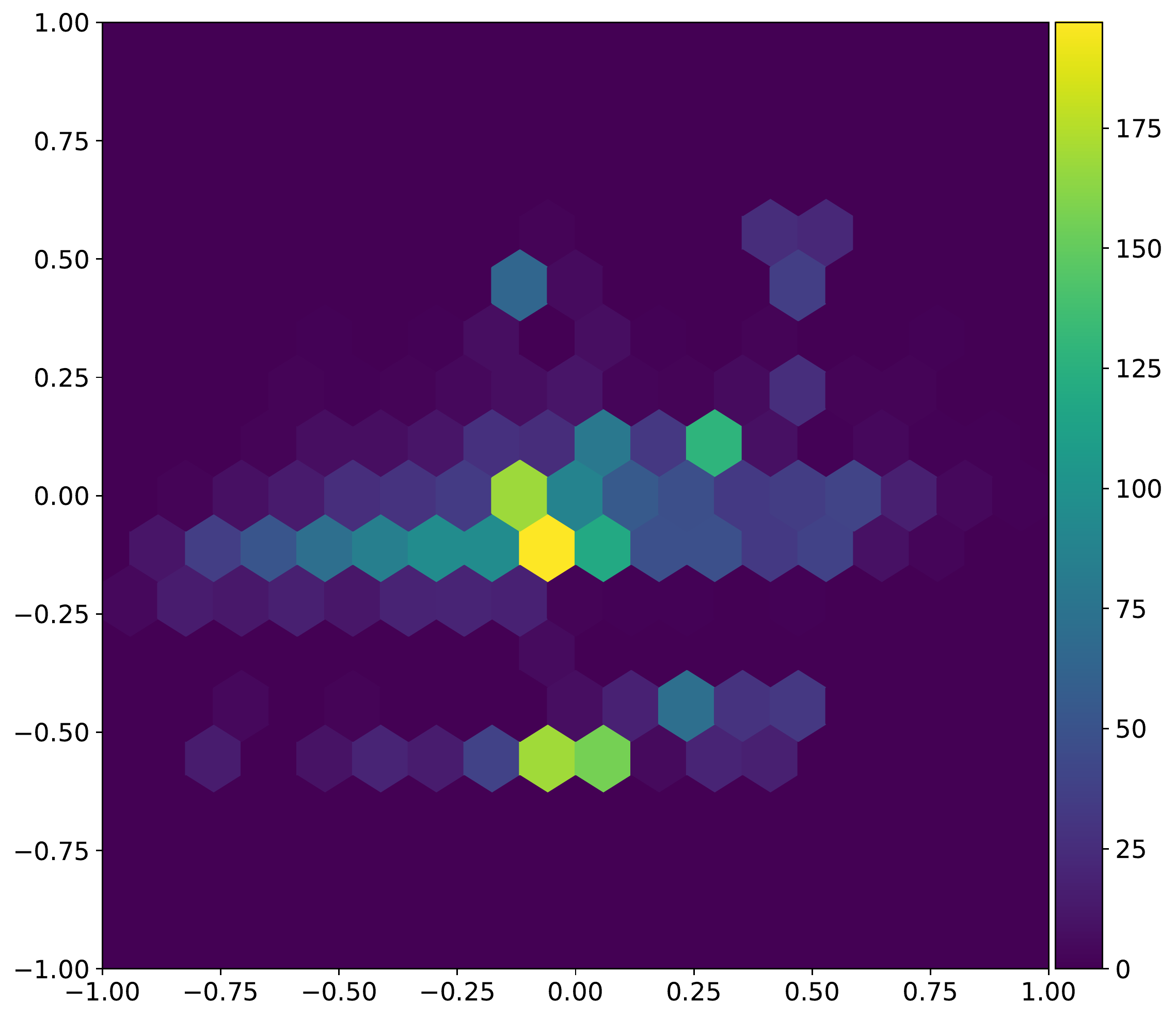}}
  \caption{\label{fig:h36m-hip}\label{fig:mpi-hip}
		{\bf Tiled 2D histogram of hip joint location} in normalized image coordinates over the test dataset for H3.6M (left) and MPI-INF-3DHP (right). The MPI-INF-3DHP set shows a wider and less regular distribution.
	}
\end{center}
\end{figure}

In Figure~\ref{fig:mpi-hip}, we can see the distribution of hip joints in the testing dataset for H3.6M and MPI-INF-3DHP. While H3.6M has most of the images around the center of the frame, MPI-INF-3DHP contains poses that are widely spread out across the image. This along with the fact that MPI-INF-3DHP uses a wider field-of-view camera explains the significant improvement we see from introducing PCL on MPI-INF-3DHP.

\begin{figure}%
\begin{center}%
\includegraphics[width=0.85\linewidth,trim={3cm 4.8cm 3cm 4.5cm},clip]{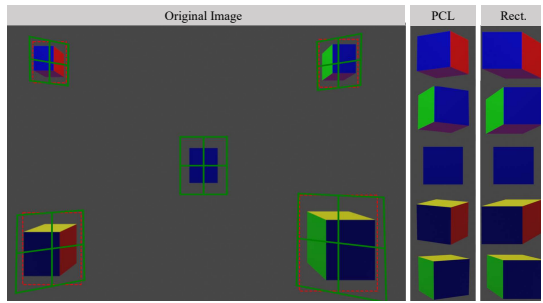}%
\end{center}%
\caption{\label{fig:cube_distortion}
		\textbf{Toy-Cube examples.} While cubes projected to the sides of an image appear distorted and look stretched when cropped, PCL undoes these perspective effects.}
\end{figure}

\section{Ablation: Model Efficiency from PCL} 

To demonstrate the effectiveness of using PCL (taking perspective distortions into account), we reduce the dimension of the hidden state of our 2D-3D keypoint lifting model and report the performance on the validation set for H3.6M. From Table~\ref{table:PCLEfficiency} we can see that even with about half the number of parameters as the baseline MLP we are able to capture a more precise reconstruction.

\begin{table}[htb]
\centering
\resizebox{\linewidth}{!}{
\begin{tabular}{l|c|c|c}
Model     & Linear Size & \# Parameters & MPJPE (mm) \\ \hline 
MLP + RC  & 1024        & 4,296,755              & 48.4       \\ \hline
\textbf{MLP + PCL} & \textbf{1024}        & \textbf{4,296,755}              & \textbf{43.8}       \\
MLP + PCL & 896         & 3,300,915              & 45.4       \\
\underline{MLP + PCL} & \underline{768}         & \underline{2,436,147}              & \underline{46.5}       \\

MLP + PCL & 512         & 1,099,827              & 50.8      
\end{tabular}}
\caption{Shown are the results on 2D-3D keypoint lifting on H3.6M while decreasing the model complexity of our PCL embedded model (MLP + PCL) while maintaining that of the baseline (MLP + RC). We can see here that despite having roughly 50\% of the parameters of the baseline, the PCL-equipped model is still able to outperform the baseline.}
\label{table:PCLEfficiency}
\end{table}

\section{Ablation: Axis-based Rotation Experiment}

To study the effects of using PCL as a post-process on exisiting trained models, we train the baseline 2D-3D keypoint lifting MLP and apply the PCL-defined rotation matrix on the predicted 3D pose. To this end, we experiment on MPI-INF-3DHP since that dataset contains more perspective distortions and will have more pronounced effects from adding the PCL-defined rotation. Along with this we also perform a "half-rotation" and "full-rotation" defined as rotation along only the x-axis and rotation about the x and y-axis respectively. Furthermore, we show results for when the 3D root is given and when scale is estimated from the 2D pose. From Table~\ref{table:PCLHalfRotation}, we can see that when we compensate the baseline model with PCL-defined rotations it improves but still falls short of our model trained end-to-end with PCL, showing the importance of incorporating PCL during model training.

\begin{table}[]
\centering
\resizebox{\linewidth}{!}{
\begin{tabular}{lll}
\multicolumn{3}{c}{MPI INF 3DHP}                                                                                    \\ \hline
\multicolumn{1}{l|}{}                           & \multicolumn{1}{c|}{2D GT + 3D Root GT} & \multicolumn{1}{c}{2D GT} \\ \hline
\hline
\multicolumn{1}{l|}{STN}                        & \multicolumn{1}{c|}{69.4}               & \multicolumn{1}{c}{74.1}       \\              
\multicolumn{1}{l|}{STN + Half Rotation (x)}    & \multicolumn{1}{c|}{67.6}               & \multicolumn{1}{c}{73.1}                      \\ 
\multicolumn{1}{l|}{STN + Full Rotation (x, y)} & \multicolumn{1}{c|}{66.3}               & \multicolumn{1}{c}{70.0}                      \\ \hline
\multicolumn{1}{l|}{PCL}                        & \multicolumn{1}{c|}{\textbf{45.6}}      & \multicolumn{1}{c}{\textbf{50.1}}             \\ \hline
\end{tabular}}
\caption{Shown are the results on 2D-3D keypoint lifting on MPI-INF-3DHP when applying half and full rotations. We can see that even after including information from PCL into the baseline model, it still falls short in terms compared to the model that is trained with PCL.}
\label{table:PCLHalfRotation}
\end{table}

\section{Ablation: Cube Dataset}

The Cube Dataset contains images of a single coloured cube with edge length 0.5m at random locations and orientations within the frame. Figure~\ref{fig:cube_distortion} shows an example of the cube used in this dataset as well as demonstrating the perspective effects that occur when objects move away from the image center. To further demonstrate the effect that ignoring perspective effects has on 3D pose estimation we introduce a variation of this dataset in which the cube has a random orientation but is always placed at the center of the frame. We refer to this variation as the Centered Cube Dataset. 
We now train models on the Centered Cube Dataset with a central crop and analyze their generalization capabilities to crops from the general dataset with the position of the cube given. On this dataset, the baseline attains an MPJPE of 13mm and our PCL variant improves to 8.2mm. The improvement for end-to-end training is equivalent, with an $4.4$mm improvement from 11.2 to 6.8mm. While this test is simplistic, the synthetic nature allows us to analyze the generalization capabilities of PCL by training on a version of the cube dataset where the cube is always centered in the training images and tested on the original version with general position, with the ground truth 2D crop location provided. PCL shows good generalization, outperforming the baseline by 36.1\% on the unseen test set. Table~\ref{table:cubeDatasetResults} compares the performance of PCL and STN models on these two datasets.
We also investigated the effect of illumination by switching from point lights to ambient illumination, which had negligible effect on the reconstruction quality.

\begin{table}[]
\resizebox{\columnwidth}{!}{%
\begin{tabular}{ll|cc|c}
                 &                    & \multicolumn{2}{c|}{Matching train-test set}    & Unseen test set \\
Model            & Detector           & centered & general &                   \\ \hline \hline
CNN + STN        & GT Loc. + GT Scale & 6.9                   & 13.0         & 13.3              \\
CNN + PCL (Ours) & GT Loc. + GT Scale & 6.9                   & \textbf{8.2} & \textbf{8.5}      \\ \hline
CNN + STN        & Trained End-to-End & 5.9                   & 11.2         & -                 \\
CNN + PCL (Ours) & Trained End-to-End & 5.9          & \textbf{6.8} & -                
\\ \hline
\end{tabular}
}
\caption{Shown are the reported MPJPE in millimeters for all tests conducted on the Cube Dataset and its variation. For this metric, lower values are better. We can see that our method produces more accurate results while at the same time generalizing better to unseen instances.}
\label{table:cubeDatasetResults}
\end{table}

\section{Implementation Details}

We normalize the 2D input poses and 3D output poses with their mean and standard deviation. On the input side, the mean and standard deviation are computed after the PCL layer and in the case of rectangular cropping (RC) after the crop and scaling operation. On the output side, the mean and standard deviation for PCL and the baselines are computed in the pelvis-centered coordinates.
We found it more effective to multiply the output by the computed standard deviation and adding the mean instead of doing the inverse operation on the label. This ensures that the network output has mean zero and unit standard deviation, which fares well with network layer initialization, while the loss operates on the scales of the original output space.

\section{Derivations of the PCL Virtual Camera}

\paragraph{Rotation Derivation.}
The rotation that maps from the virtual to the real camera, $\mRvr$, stems from the definition of rotation matrices. We use the right-handed rule, i.e., counter-clockwise rotation for positive angles and a right-handed coordinate system with the y-axis pointing downwards, x-axis rightwards, and positive z pointing in camera direction. The definition of $\mRvr = \mR_{y} \mR_x$ reads thereby
\begin{align}
\mR_{y} & \mR_x = 
\left[ \footnotesize {\begin{array}{ccc}
		\cos(\phi) & 0 & \sin(\phi)\\
		0 & 1 & 0\\
		-\sin(\phi) & 0 & \cos(\phi)\\
\end{array} } \right]\hspace{-0.1cm}
\left[ \footnotesize {\begin{array}{ccc}1 & 0 & 0\\
	0 & \cos(\theta) & -\sin(\theta)\\
	0 & \sin(\theta) & \cos(\theta)\\
	\end{array} } \right]\nonumber\\
& =
\left[ \footnotesize {\begin{array}{ccc}
		\cos(\phi) & \sin(\phi)\sin(\theta) & \cos(\theta)\sin(\phi)\\
		0 & \cos(\theta) & -\sin(\theta)\\
		-\sin(\phi) & \sin(\theta)\cos(\phi) & \cos(\phi)\cos(\theta)\\
\end{array} } \right]\;,
\label{eq:eulermat}
\end{align}
where $\phi$ and $\theta$ are, respectively, the vertical and horizontal rotations depicted in Figure~2 of the main document. The equation given in the main document follows from the following trigonometric relations,
\begin{align}
\sin(\phi) &= \frac{{\vp}_x}{\sqrt{1+{\vp}_x^2}},\;
&\cos(\phi) &= \frac{1}{\sqrt{1+{\vp}_x^2}},\;\nonumber\\
\sin(\theta) &= \frac{-{\vp}_y}{\sqrt{1+{\vp}_x^2+{\vp}_y^2}},\;
&\cos(\theta) &= \frac{\sqrt{1+\vp_x^2}}{\sqrt{1+{\vp}_x^2+{\vp}_y^2}},
\label{eq:trigon}
\end{align}
where $\vp$ is the point on the original image plane to which the virtual camera is rotated. 

\paragraph{Virtual Focal Length Selection}

\begin{figure}[t]
\begin{center}
    \includegraphics[width=1\linewidth,trim={3.5cm 3cm 3.5cm 3cm},clip]{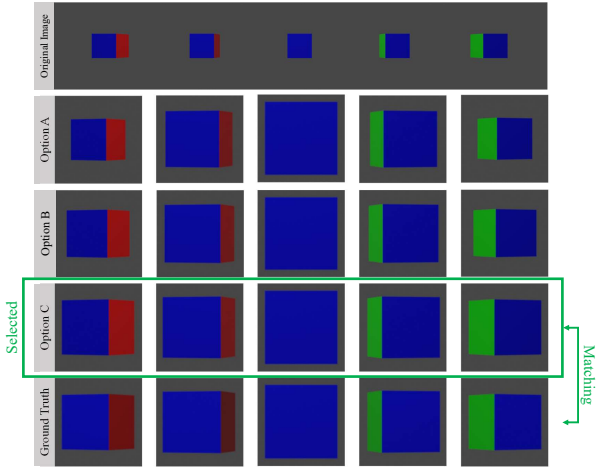}
\end{center}
\caption{\label{fig:pcl_focal}
		\textbf{Effect of virtual camera focal length.} The proposed options for setting the virtual focal length scale differently with respect to the image position. When given the pixel cube width as input (as a factor of the entire image resolution), only option C maintains the desired margin of 10 \% between cube and crop boundary. The ground truth is a cube placed at the center of the screen rotated by the same angle that the virtual camera is rotated by. Remaining differences in color stem from the position-dependent illumination effects.
	}
\end{figure}

The focal length of the virtual camera defines the zoom level of the PCL crop. It controls the crop size in the original images. Therefore, it needs to be set individually for every crop target, depending on its desired size and position in the image. In the following, we explain the derivation of the three options we propose. Figure~\ref{fig:pcl_focal} shows example crops of each method and their tightness of fit.

\begin{itemize}
	\item[\textbf{A.}] By setting $\mathbf{\vh^\text{virt}}$ to $\mathbf{f}$, the camera is only rotated, without any change in zoom. A crop is obtained by scaling with factor $\vs$, that means, $\vf = \frac{\vh}{\vs}$. Figure~\ref{fig:pcl_focal}, second row, shows that this simple choice leads to inconsistent crop sizes. The object appears smaller the further away it is.
    \item[\textbf{B.}] By multiplying the virtual length with $\|\vp\|$, the distance of the crop target position on the image plane to the camera center, this distance-related effect is compensated. However, as the third row in Figure~\ref{fig:pcl_focal} shows, this match is not perfect as it does not account for the foreshortening effect when projecting from the original image plane onto the virtual one.
	\item[\textbf{C.}] Our final choice models foreshortening with\\
	$\vh^\text{virt}_x = \vf_x  \|{\vp}\| \sqrt{\vp_x^2+1}$ and\\
	$\vh^\text{virt}_y = \vf_y \frac{\|{\vp}\|^2}{\sqrt{\vp_x^2+1}}$. It is derived as follows.
\end{itemize} 

\paragraph{Derivation of Option C.} Let $\vp = (x,y,z)^\top$ be the target crop position on the image plane, a 3D position. By construction, $\vp$ will be at the image center. Therefore, projecting the infinitesimal motion offset $\vp+(\delta x, \delta y, 0)^\top$ and comparing the ratio of the offset in the original and projection yields the desired scale estimate. Formally, we write
\begin{equation}
    (u,v,1)^T = \mP \left( \vp+(\delta x, \delta y, 0)^\top \right),
\end{equation}
where $\mP$ projects points in the original coordinate system to the virtual one, as defined in the main document. For the sake of simpler equations we do computations in camera coordinates with the origin at the image center and the focal length $\vf=1$. In this case, $\mP=\mRrv$. Using the definition of $\mRvr$ above, the identity $\mRrv = \mRvr^\top$, and computing the partial derivatives with respect to $\delta x$ and $\delta y$ at $\delta x = \delta y =0$ we obtain
\comment{
\begin{equation}
\frac{\partial (u,v,1)^T}{\partial \delta x} = \frac{(\sqrt{1 + y^2/(1 + x^2)} (1 + x^2 + y (\delta y + y))}{(1 + x (\delta x + x) + y (\delta y + y))^2}.
\end{equation}
Evaluated at $\delta x = 0$, this gives
}
\begin{equation}
\frac{\partial (u,v,1)^T}{\partial u} \Bigr\rvert_{\delta u = 0} = \frac{1}{\sqrt{(1 + x^2)} \| \vp \|}
\end{equation}
and
\begin{equation}
\frac{\partial (u,v,1)^T}{\partial v} \Bigr\rvert_{\delta v = 0} = \frac{\sqrt{(1 + x^2)}}{\| \vp \|^2}.
\end{equation}
These equations compute the horizontal and vertical pixel scale ratio between the original and virtual image at $\vp$. To maintain the scale, the focal length must be set to the inverse of this scaling factor, which is Option C for $f^\text{virt}$ that also incorporates the original focal length and the desired crop scale.

Note that the equations for $\vh^\text{virt}_x$ and $\vh^\text{virt}_y$ are not equivalent with $x$ and $y$ exchanged because the $x$ and $y$ axes in the virtual camera do not, in general position, project to perpendicular lines in the original view. Figure~1 of the main paper provides examples of this perspective effect. In our definition, the up-direction is kept fixed and the horizontal axis is rotated, therefore, behaving differently in the slant compensation. An equivalent formulation could be derived with the horizontal axis fixed and the vertical axis rotated.

\paragraph{Maintaining the original aspect ratio}
To enable computations with a different focal length in the $x$ and $y$ directions, which models the case of non-square pixels, the above notation was performed individually for the horizontal and vertical directions. This, however, can lead to stretched crops if different scales are predicted for $\vs$ in horizontal and vertical directions. To maintain the original aspect ratio, we set the focal length to the minimum of the axis-specific lengths. This leads to a crop that is the same or larger than the original one with mismatching aspect ratio, thereby strictly containing the object of interest.

{\small
	\bibliographystyle{ieee_fullname}
	\bibliography{bib/string,bib/vision,bib/learning,bib/biomed}
}

\end{document}